\documentclass[format=sigconf]{acmart} 

\AtBeginDocument{%
 }

\copyrightyear{2024}
\acmYear{2024}
\setcopyright{rightsretained}
\acmConference[GECCO '24 Companion]{Genetic and Evolutionary Computation Conference}{July 14--18, 2024}{Melbourne, VIC, Australia}
\acmBooktitle{Genetic and Evolutionary Computation Conference (GECCO '24 Companion), July 14--18, 2024, Melbourne, VIC, Australia}
\acmDOI{10.1145/3638530.3664138}
\acmISBN{979-8-4007-0495-6/24/07}

\usepackage{enumitem}
\usepackage{multirow}
\usepackage{graphicx}
\usepackage{subcaption} 
\usepackage{algorithm}
\usepackage[commentColor=teal,italicComments=false]{algpseudocodex}

\usepackage{stackengine}
\setstackEOL{\\}

\algnewcommand\algorithmicforeach{\textbf{for each}}
\algdef{S}[FOR]{ForEach}[1]{\algorithmicforeach\ #1\ \algorithmicdo}

\begin{document}

\title{The Effect of Training Schedules on Morphological Robustness and Generalization}

\author{Edoardo Barba}
\affiliation{%
 \institution{University of Trento}
 \streetaddress{Via Calepina, 14}
 \city{Trento}
 \country{Italy}
 \postcode{38122}
}

\author{Anil Yaman}
\affiliation{%
 \institution{Vrije Universiteit Amsterdam}
 \streetaddress{De Boelelaan 1105}
 \city{Amsterdam}
 \country{Netherlands}
}

\author{Giovanni Iacca}
\affiliation{%
 \institution{University of Trento}
 \streetaddress{Via Calepina, 14}
 \city{Trento}
 \country{Italy}
 \postcode{38122}
}


\begin{abstract}
Robustness and generalizability are the key properties of artificial neural network (ANN)-based controllers for maintaining a reliable performance in case of changes. It is demonstrated that exposing the ANNs to variations during training processes can improve their robustness and generalization capabilities. However, the way in which this variation is introduced can have a significant impact. In this paper, we define various training schedules to specify how these variations are introduced during an evolutionary learning process. In particular, we focus on morphological robustness and generalizability concerned with finding an ANN-based controller that can provide sufficient performance on a range of physical variations. Then, we perform an extensive analysis of the effect of these training schedules on morphological generalization. Furthermore, we formalize the process of training sample selection (i.e., morphological variations) to improve generalization as a reinforcement learning problem. Overall, our results provide deeper insights into the role of variability and the ways of enhancing the generalization property of evolved ANN-based controllers.
\end{abstract}

\ccsdesc[500]{Theory of computation~Evolutionary algorithms}
\ccsdesc[500]{Computing methodologies~Neural networks}

\keywords{Neuroevolution, Generalization, Robustness, Variability}

\maketitle


\section{Introduction}
\label{sec:intro}

The robustness and generalizability of artificial neural network (ANN)-based controllers are crucial properties for providing reliable performance when facing changes. The physical world is, in fact, highly uncertain and is subject to variations of various kinds. To operate effectively, controllers must be robust to these variations, meaning they should be able to perform effectively in diverse conditions. 

Robustness is typically defined as the ability to maintain sufficient performance in case of small perturbations relative to the training data. Whereas, generalization is the ability to handle a wider range of differences, especially concerning scenarios that are not encountered during the training process \cite{mangal2019}. 

For instance, in the case of a walking robot, an ANN-based controller should be robust to varying terrains, obstacles, or changes to the robot morphology, e.g. due to damages. A controller trained to handle only one specific morphology or one specific environment would reduce its robustness and generalizability, making it more fragile to perturbations \cite{triebold2023evolving, putter2017}. 

Robustness and generalizability can also be vital for reducing the simulation-to-reality gap (reality gap for short), which refers to the mismatch between simulated and real-world environments \cite{salvato2021crossing}. The training conditions in simulation may not capture the noise and unexpected situations that can be encountered in the real world. While optimizing for \emph{all} possible unexpected situations may not even be possible (because such situations are most likely unknown), it is plausible to think that more generalizable controllers should handle such situations better than highly specialized controllers.

Because of the importance of these properties, there is a growing interest in developing controllers that exhibit robustness across a broad spectrum of morphological variations and can generalize to unseen morphologies \cite{Carvalho2023}. The main approaches to handling unexpected situations are retraining, relying on repertoires \citep{cully2015robots}, or online adaptation \citep{yaman2021evolving, nygaard2021real, risi2013confronting}. However, these approaches can be costly since they rely on a form of learning or memorizing a repertoire. On the other hand, a generalist controller may be able to handle a large range of variability without the need for adaptation \cite{triebold2023evolving}.


In various domains, it has been observed that exposing learners to more variability can be an effective strategy for enhancing their robustness and generalization capabilities \cite{RAVIV2022462}. 
More variability tends to make initial learning more challenging but eventually leads to more general and robust performance. This core principle has been discovered in many domains (e.g., visual perception \cite{HUSSAIN201289}, motor learning \cite{huet2011, schmidt1997}, and language acquisition \cite{posner1968}). In supervised learning tasks, training variability can be enhanced, for instance, through techniques like data augmentation \cite{gliozzi2019grounding, hill2019environmental, hernandez2018}.
More variability during training is therefore essential to generalize learning, as it facilitates the formation of more abstract knowledge \cite{posner1968}. Variability is also essential for an ``antifragile'' system, a concept introduced by Taleb \cite{Taleb12} to describe systems that can benefit from exposure to variations. Taleb distinguishes between fragile, robust, and antifragile systems, asserting that complex systems fall into the antifragile category due to their opaque nature. ANNs, with their nonlinear relationships, can benefit from environmental variations, making adaptive controllers antifragile as they navigate uncertainties. Introducing too much variability during training, however, can be counterproductive, as high variability can make learning more difficult when learners are in the very early stages of acquiring a target behavior \cite{RAVIV2022462}. It is therefore crucial to understand the best way of introducing variability to improve generalization. 

One approach to improve generalization by introducing variability is presented in \cite{triebold2023evolving}, where the authors employ Evolution Strategies to optimize ANNs to control different body shapes in locomotion tasks. They enable generalization by introducing a different morphology (generated by modifying body parameters such as leg length or width) at each generation and use it as the evaluation fitness for that generation. 

The present work builds upon a similar idea to study the effects of different training schedules used to evolve the ANN in depth. Our main focus is to investigate how variability influences the learning and generalization of evolved ANN-based controllers.
To this aim, we perform a thorough analysis using three OpenAI Gym tasks: Bipedal Walker, Walker2D, and Ant. The training methodologies follow the algorithm presented in \cite{triebold2023evolving}, introducing morphological variations in each generation. However, differently from \cite{triebold2023evolving}, here we specifically focus on a thorough study of different training schedules. 

In our analysis, we find that increasing the diversity of morphologies during training improves both robustness and generalization. Sampling variations from a continuous domain (i.e., modifying various morphological parameters) during training provides more robustness than a discrete set of variation possibilities. Moreover, using the Beta distribution for sampling the variations for training enhances generalization. We also confirm the observation reported in \cite{triebold2023evolving}, that the order in which the morphological variations are introduced during training significantly impacts generalization performance. Finally, we formalize the process of online learning of training sample selection as a reinforcement learning problem and show its learning dynamics. In particular, we study if some samples (i.e., morphological variations) are preferred, and how this impacts the final generalization. Taken together, these results highlight the importance of introducing variation during training and its effects on the generalization capabilities of ANN-based controllers.

The rest of the paper is organized as follows. The next section presents the background and related works. Section \ref{sec:methods} describes our methods. Then, Section \ref{sec:setup} and Section \ref{sec:results} present the experimental setup and the numerical results. Finally, Section \ref{sec:concl} concludes this work.


\section{Background and Related Works}
\label{sec:rw}

\subsubsection*{Neuroevolution through Evolution Strategies.}

ANN-based controllers are used in a wide range of control tasks in simulation and the real world. Often these tasks require black-box optimization approaches due to the difficulty of obtaining a closed-form objective function for which gradient-based approaches can be used. Neuro-evolution is a derivative-free approach that is used to optimize the weights (and, potentially, the structure) of ANNs using evolutionary algorithms \cite{floreano2008neuroevolution, stanley2019}. Due to the large dimensionality of the ANN search space, however, neuro-evolutionary approaches typically suffer from scalability issues. In this regard, Evolution Strategies (ES) have been shown to scale particularly well, due to their capability of estimating the gradient based on sampling \cite{Wierstra2011,salimans2017}. Therefore, ES has been applied on a wide range of control tasks that require optimization in large dimensional parameter spaces \cite{pagliuca2020efficacy,todorov2012}. 

\subsubsection*{Morphological Robustness and Generalizability}

As outlined by Raviv et al. \cite{RAVIV2022462}, variability has been demonstrated to be a crucial aspect in improving the generalization capabilities of learners. In the real world, one can say that learning intrinsically requires some form of generalization. The prevailing relation between the effects of variability and generalization, observed across various domains, suggests that learning from less variable input tends to occur rapidly but may struggle to generalize to new stimuli. Conversely, learning from more varied input typically starts slower but often results in superior generalization.

Variability, however, has many dimensions. The same authors of \cite{RAVIV2022462} identified four distinct types: Numerosity (more or fewer examples), Heterogeneity (similarity between samples), Situational Diversity (more or less variable conditions), and Scheduling (different order of samples). The latter, in particular, is also related to the timing of variability, which can directly impact the learning process. It has been shown that the impact of variability can vary significantly, depending on the stage of the learning process \cite{RAVIV2022462, boyce2006implications}. 

The common intuition is that excessive variability during the initial stages of learning can hinder learning. This is because, when individuals are just starting to acquire a new skill or behavior, they may encounter difficulties if they are exposed to too much variability too soon \cite{magill1990review, hebert1996practice}. Instead, beginning with simpler and less variable examples tends to facilitate learning. This concept is the foundation of curriculum learning, a methodology inspired by these learning patterns where skills are progressively built upon simpler concepts \cite{curr_learning_2021}. Eventually, exposing a learner to variability is however necessary to improve its ability to generalize learning to new contexts, as it facilitates the formation of more abstract knowledge. However, deciding when to expose the learner to variability is not trivial. In the following, we will refer to the way to decide which variations a learner should be exposed to, and when, with the term \emph{training schedule}.

A preliminary study on training schedules was conducted in \cite{triebold2023evolving}, aiming to improve the generalization of ANN-based controllers by exposing them to variations in each generation of an evolutionary process. Focusing in particular on the case of morphological variations, the authors showed that this approach provides better generalizability performance relative to a baseline when the average performance on all morphological variations is maximized. 

In the case of morphological variations, finding ways to achieve robustness can help maintain the performance in case of unexpected changes in the physical structure of the agent without requiring retraining \cite{triebold2023evolving}. We refer to \emph{morphological robustness} as the ability of a controller to maintain stable performance despite small perturbations (w.r.t. the training scenarios \cite{putter2017}) in the morphological parameters of the agent (e.g., a robot) controlled by that controller. Whereas, \emph{morphological generalizability} refers to the ability to maintain a sufficient performance in case of larger changes in the morphological parameters of the agents \cite{Rajeswaran2017Generalization}.


\section{Methods}
\label{sec:methods}


The training process of our approach is illustrated in Algorithm \ref{alg:tranining}. We use a similar approach proposed by \cite{triebold2023evolving}, as we define a set of morphologies that are introduced during the evolutionary process. These morphologies are concerned with the parameter ranges of the physical structure of the agents. Thus, we consider a training set $M = \{m_1, m_2, \dots, m_n\}$ that is used during training, and a validation set $V = \{v_1, v_2, \dots, v_n\}$ that is used for validating the generalization score. $M$ and $V$ are initialized or sampled from the same morphological parameter ranges, but they are used differently, because $V$ consists of a discrete and predefined set, whereas $M$ may be defined before the evolutionary run, or can be generated automatically during the process. We then have a training schedule $T$ that determines how the morphologies in the training set $M$ are introduced during the evolutionary process.

\begin{algorithm}[ht!]
\caption{Training Process}
\label{alg:tranining}
\begin{algorithmic}[1]
\State Initialize $G$ as an empty set \Comment{Generalist set}
\State Initialize $M$ as the training set of morphologies
\State Initialize $T$ as the used training schedule
\State Initialize $V$ as the validation set of morphologies
\Function{EvolutionaryProcess}{$M$, $T$, $V$}
    \State $I_{\text{best}} \gets \{\}$
    \State $ f \gets \infty$ \Comment{Assuming minimization}
    \State $I \gets \text{initialize}()$ \Comment{Initialize an individual} 
    \For{$i < \text{maxGenerations}$}
        \State $m_i \gets \Psi (M, T, i)$ \Comment{Choose $m_i$ from $M$ based on $T$ and $i$}
        \State $f_{m_i} \gets \text{evaluate}(I, m_i)$ \Comment{\Longunderstack[r]{Sample solutions from $I$ and\\evaluate them on $m_i$}} 
        \State $I_{\text{best}}$ = individual with highest fitness on $m_i$
        \State $I \gets \text{updateParameters}(I,f_{m_i})$ \Comment{\Longunderstack[r]{Update $I$ based\\on fitness}} 
        \For{\textbf{each} $v$ \textbf{in} $V$}
            \State $f_v \gets \text{evaluate}(I_{\text{best}},v$)
        \EndFor
        \State $f_{\text{best}}$ = mean fitness of $I_{\text{best}}$ on all $v$ in $V$
        \If{$f_{\text{best}} < f$} \Comment{Assuming minimization}
            \State $G\gets \text{add}(I_\text{best})$ \Comment{Add to the generalist set}
            \State $f \gets f_{\text{best}}$
        \EndIf
    \EndFor
\EndFunction
\end{algorithmic}
\end{algorithm}

The training process begins with an empty ``generalist set'' $G$, followed by the generation of the initial population, where each individual represents the weights of the ANN. 
Then, at each generation $i$, the population is evaluated based on the fitness value obtained on a morphology $m_i$. Once each individual is evaluated, the one with the best-obtained score is selected, and its average fitness ($f_{\text{best}}$) across all morphologies in $V$ is computed. The performance of a controller $I_{\text{best}}$ on a set of different morphologies $V$ is defined as the average performance evaluated on all morphologies in $V$. This is mathematically expressed as:
\begin{equation}
f_\text{best} = \frac{1}{|V|} \sum_{j=1}^{|V|} \text{evaluate}(I_\text{best}, v_j)
\end{equation}
where $f_\text{best}$ represents the average performance of the controller $I_\text{best}$ over morphologies in $V$, $|V|$ denotes the total number of morphologies in set $V$, and $\text{evaluate}(I_\text{best},v_j)$ is the evaluation function that returns the score of the controller $I_\text{best}$ on morphology $v_j$. 

During the evolution process, every time a controller achieves a better average fitness score on the validation set $V$, this controller is added to the generalist set $G$. Therefore, at the end of the process, the last controller added has the highest average fitness score on $V$. Note that in the algorithm, the $\Psi$ function determines which morphology to select at generation $t$ based on the type of the training schedule $T$.

For the evolutionary process, we use the eXponential Natural Evolution Strategy (xNES) algorithm \cite{glasmachers2010exponential} available in the EvoTorch implementation \cite{toklu2023evotorch}\footnote{The code of our algorithm is publicly accessible at: \url{https://github.com/edoardobarba/evolving-generalist-controllers}}.

\subsection{Training Schedules}
To examine the impact of various ways of introducing variability during the evolutionary process, we define several training schedules. We refer the reader to the Supplementary Material for visualization and additional information regarding these training schedules, which are detailed next.



\subsubsection*{Discrete Random}
This training schedule serves as a baseline. It involves a training set comprising $36$ morphologies, generated with parameters with equal intervals within the possible morphological parameter space. At each generation, the morphology is chosen \textit{randomly with uniform probability} from one of these morphologies.

\subsubsection*{Discrete Incremental}
The discrete incremental training schedule uses the same morphology set as the discrete random but it selects the morphological variations in an incremental order on 2-dimensional parameter space starting from the initial x and y morphological parameters and incrementing the parameters along the x-axis first, followed by the increments on the parameters along the y-axis. Morphologies therefore are defined first with small body parameters (e.g., leg length) and then such parameters gradually increase. When the maximum parameter for x and y is reached, the process returns to the initial state. Defining morphologies in this way ensures a gradual transition between morphologies, minimizing large leaps from one morphology to another, and thus facilitating the learning process. 


\subsubsection*{Uniform}
This schedule samples morphology parameters randomly with uniform probability from the 2-dimensional continuous space at each generation. This differs from the discrete case in providing more variability. 

\subsubsection*{Gaussian}
Gaussian training schedule uses a Gaussian distribution that is centered in the center of the 2-dimensional morphology parameter space with $\Sigma = \big(\begin{smallmatrix}
  \sigma^2 & 0\\
  0 & \sigma^2
\end{smallmatrix}\big)$. The rationale behind this choice is to moderate the extent of variation introduced during the sampling focused in the center of the morphology space. 


\subsubsection*{Cauchy}
To introduce more variability, but still centered around the center of the morphology space, we used the Cauchy distribution. By leveraging a fat-tailed distribution like Cauchy, there is more probability of selection of variation far from the center.

\subsubsection*{Beta}
We use the Beta distribution to sample the morphology parameters with $\alpha$ and $\beta$ parameters set both to $0.1$. This schedule primarily samples from the outer edges of the parameter space, resulting in significant variability during training. Each generation encounters markedly different morphologies, potentially leading to diverse behaviors required for effective control. This strategy aims to enhance generalization capability by exposing the trained ANN to a wide range of morphologies.

\subsection{Multi-armed Bandit Approach}

This section presents a method to learn to dynamically select training morphologies during the evolutionary process. Indeed, certain morphologies may be more useful than others for improving the generalizability of the controllers. Thus, to achieve that, we formalize the learning process as a reinforcement learning approach on a Multi-Armed Bandit (MAB) problem \cite{bubeck2012regret, raj2017taming, guo2019autosem}. Based on this formalization, we can keep a utility estimate of each morphological variation. Then, we can make a selection based on the utilities. After this selection and evaluation, we can update the estimate of the utility of the selected variation based on its reward, that is whether or not this selection provided an improvement in generalization. 

\subsubsection*{Formalization}
We have a $N$ number of arms $\{ a_1, \ldots, a_N \}$ corresponding to the set of morphological variations $\{ D_1, \ldots, D_N \}$. At each generation, we choose a morphology to evaluate, and subsequently update its utility based on the observed reward $r$. We characterize the utility of each morphological variation $k$ with a value $\theta_k$ ranging between $0$ and $1$.
The agent initially holds a prior probability distribution, which is subsequently updated. The distribution is represented by Beta distributions, characterized by parameters $\alpha_k$ and $\beta_k$. Therefore each morphology $k$ is associated with its own $\alpha_k$ and $\beta_k$ parameters. The probability density function is given by:
\begin{equation}
f(\theta_k | \alpha_k, \beta_k) = \frac{1}{B(\alpha_k, \beta_k)} \theta_k^{\alpha_k-1} (1-\theta_k)^{\beta_k-1}.
\end{equation}

\subsubsection*{Reward and Update}
After each generation $t$ of the training process, the $\alpha$ and $\beta$ parameters of all morphologies are updated based on a reward.
The generalization capability $f$, defined as the average score across all morphologies in the validation set, serves as the basis for determining the reward. Specifically, a reward $r_{t}$ is granted if the chosen morphology yields an improvement in the generalization capability compared to the moving average of the capability over the preceding $s$ generations:
\begin{equation}
r_{t} = 
\begin{cases} 
1, & \text{if } f_{t} > \text{average of } f \text{ over the last } s \text{ generations} \\
0, & \text{otherwise}
\end{cases}
\end{equation}
where $s$ represents a hyper-parameter, set to $10$ for our experiments.



We can then leverage the ``conjugacy'' property of the Beta distribution, which states that the posterior distribution of a Beta distribution is also a Beta distribution. This property simplifies the updating process of the parameters $\alpha_k$ and $\beta_k$ after each training iteration. Specifically, after each generation, the parameters $\alpha_k$ and $\beta_k$ of all the morphologies in the validation set can be updated with the following rule \cite{russo2018}:
\begin{equation}
(\alpha_k, \beta_k) = 
\begin{cases} 
(\hat{\alpha}_k, \hat{\beta}_k), & \text{if } k \neq x^{t} \\
(\hat{\alpha}_k, \hat{\beta}_k) + (r_{tb}, 1 - r_{tb}), & \text{if } k = x^{t}
\end{cases}
\end{equation}
where 
\[
\hat{\alpha}_k = (1-\gamma)\alpha_k + \gamma\alpha_0 
\quad \text{and} \quad 
\hat{\beta}_k = (1-\gamma)\beta_k + \gamma\beta_0,
\]
and $\gamma$ controls how quickly uncertainty is injected into the system. Moreover, $x^{t}$ denotes the morphology chosen at time step $t$ (i.e., at generation $t$), whereas $\alpha_0$ and $\beta_0$ represent the prior probabilities. 
Since the utility of each morphology varies during the training process (e.g., some morphologies can be more useful at the beginning of the training process while others at the end), we can consider the system as non-stationary. To address this system property, inspired by the methodology proposed in \cite{guo2019autosem}, one simple approach involves diminishing the significance of past observations at each step with the introduction of a tunable decay ratio $\gamma$.


\subsubsection*{Morphological Variation Choice}
At each generation, the choice of which morphology to use relies on Thompson Sampling \cite{russo2018}, namely, We choose the morphology corresponding to the highest $\theta^k$ value (i.e., $\text{argmax}_k \theta^k$). In essence, Thompson Sampling introduces a balance between exploration (trying out new morphologies) and exploitation (sticking with the best-known ones), allowing the model to learn and adapt based on the outcomes of its actions, leading to more informed and effective decisions over time.


\section{Experimental Setup}
\label{sec:setup}

\subsubsection*{Environments}
Experiments were conducted in various OpenAI Gym continuous control problems \cite{gym2016}, typically employed for testing reinforcement learning algorithms. Specifically, we employed the \textit{Bipedal Walker} environment from \textit{Box2D} \cite{catto2011box2d}, along with \textit{Walker2D} and \textit{Ant} from \textit{MuJoCo} \cite{todorov2012}. In each of these environments, two body parameters were varied to train and test the ANNs on diverse morphologies. These body parameters are leg width and leg length for Bipedal Walker, and upper leg length and lower leg length for both Walker2D and Ant. The parameter ranges for the training and test sets differ across control tasks and are reported in Table \ref{tab:evaluation_ranges}. 

\subsubsection*{The ANN-based Controllers}
The controller used in this work is a fully connected feedforward ANN with the same topology across all tasks. This topology comprises input, hidden, and output layers, with the variation among tasks lying in the number of neurons. The network structures utilized in this paper are derived from \cite{triebold2023evolving}, which, in turn, draws from prior investigations \cite{pagliuca2020efficacy}. The hidden layer comprises 20 neurons, including one bias neuron, across all three tasks. The input layer consists of 24 neurons for the Bipedal Walker task, 17 neurons for the Walker2D task, and 27 neurons for the Ant task, respectively. Conversely, the output layer consists of 4 neurons for the Bipedal Walker task, 6 neurons for the Walker2D task, and 8 neurons for the Ant task, respectively. Finally, the tangent hyperbolic (tanh) activation function is employed for all neurons in the network.

\begin{table*}[ht!]
 \centering
 \caption{Ranges of parameters for each task.} 
 \label{tab:evaluation_ranges}
 \begin{tabular}{llccc}
 \toprule
 \textbf{Task} & \textbf{Parameter Set} & \textbf{Training set} & \textbf{Testing Set} & \textbf{Step} \\
 \midrule
 \multirow{2}{*}{Bipedal walker} & Leg Length & [7, 17] & [3, 6] $\cup$ [18, 21] & 2 \\
 & Leg Width & [24, 44] & [16, 23] $\cup$ [45, 52] & 4 \\
 \midrule
 \multirow{2}{*}{Walker2d} & Lower leg length & [0.3, 0.425] & [0.225, 0.25] $\cup$ [0.45, 0.5] & 0.025 \\
 & Upper leg length & [0.4, 0.65] & [0.25, 0.35] $\cup$ [0.7, 0.8] & 0.05 \\
 \midrule
 \multirow{2}{*}{Ant} & Lower leg length & [0.5, 1.5] & [0.2, 0.4] $\cup$ [1.6, 1.9] & 0.1 \\
 & Upper leg length & [0.7, 1.7] & [0.4, 0.6] $\cup$ [1.8, 2.1] & 0.1 \\
 \bottomrule
 \end{tabular}
\end{table*}

\subsubsection*{Validation Sets}
A validation set is utilized to test the generalizability of controllers during training. For all three tasks, the validation set encompasses $36$ morphologies, with the parameter values within the ranges outlined in Table \ref{tab:evaluation_ranges}, under the ``Training set'' column. For visual representations of the validation sets for the three tasks, please consult the Supplementary Material.

\subsubsection*{Experimental parameters}
For the Bipedal Walker task, the experiment employs 3000 generations, with each episode consisting of 1000 steps. For the Walker2D and Ant tasks, the number of generations is set to 5000, with each episode comprising 1000 steps. 

\subsubsection*{Evaluation Metrics} \label{sec:testing}
The evaluation phase aims to measure the robustness and generalization capabilities of the ANN. As mentioned, we refer to robustness as the ability to control morphologies within the parameter ranges of the training set, while to generalization as the ability to control morphologies outside these parameter ranges, which have not been encountered during the training process. The evaluation approach draws inspiration from the methodology introduced by Packer et al. \cite{packer2018assessing} for assessing the generalization of deep reinforcement learning controllers. Namely, we test robustness to in-distribution samples (interpolation) and generalization to out-of-distribution samples (extrapolation). Interpolation implies that agents should perform well in controlling morphologies where parameters are similar to those seen during training. Extrapolation requires agents to perform well in morphologies where parameters are different from those seen during training. We utilize two distinct evaluation sets to address these aspects:
\begin{enumerate}[leftmargin=*]
 \item \textbf{Training morphologies}: a set of morphologies with the same parameter ranges used for training allows us to assess the robustness of the controller on \textit{in-distribution samples}.
 \item \textbf{Testing morphologies}: a set of morphologies with parameters ranges outside those used for training, to evaluate the ability to generalize on \textit{out-of-distribution samples}. 
\end{enumerate}

We can therefore define the average performance obtained on morphologies on the training set as a measure of \textit{robustness} within the context of the data the controller has been exposed to during training. Meanwhile, the average scores on morphologies on the test set provide insight into its ability to \textit{generalize} beyond the data it has seen during training. For a visual representation of the training and testing sets, refer to the Supplementary Material.

\subsubsection*{Statistical Analysis}
The results presented in the following Section represent the median values obtained from $30$ independent evolutionary training runs. To determine the statistical significance of the observed differences, the Mann-Whitney U test \cite{mann1947test} was utilized with confidence level $\alpha=0.05$. In the figures presenting the results, the statistical test outcomes are depicted using bars with red asterisks placed above the box plots, denoting groups that exhibit significant differences.


\section{Results}
\label{sec:results}

We present our results separately analyzing first the effect of the heterogeneity and of the scheduling of the training morphologies. Then, we present the results of the MAB approach. Finally, we analyze the distribution of morphologies selected during training.

\subsection{Effect of the Heterogeneity of Morphologies}
\begin{figure}[ht!]
 \includegraphics[clip, trim=0cm 0cm 0cm 1.2cm, width=\linewidth]{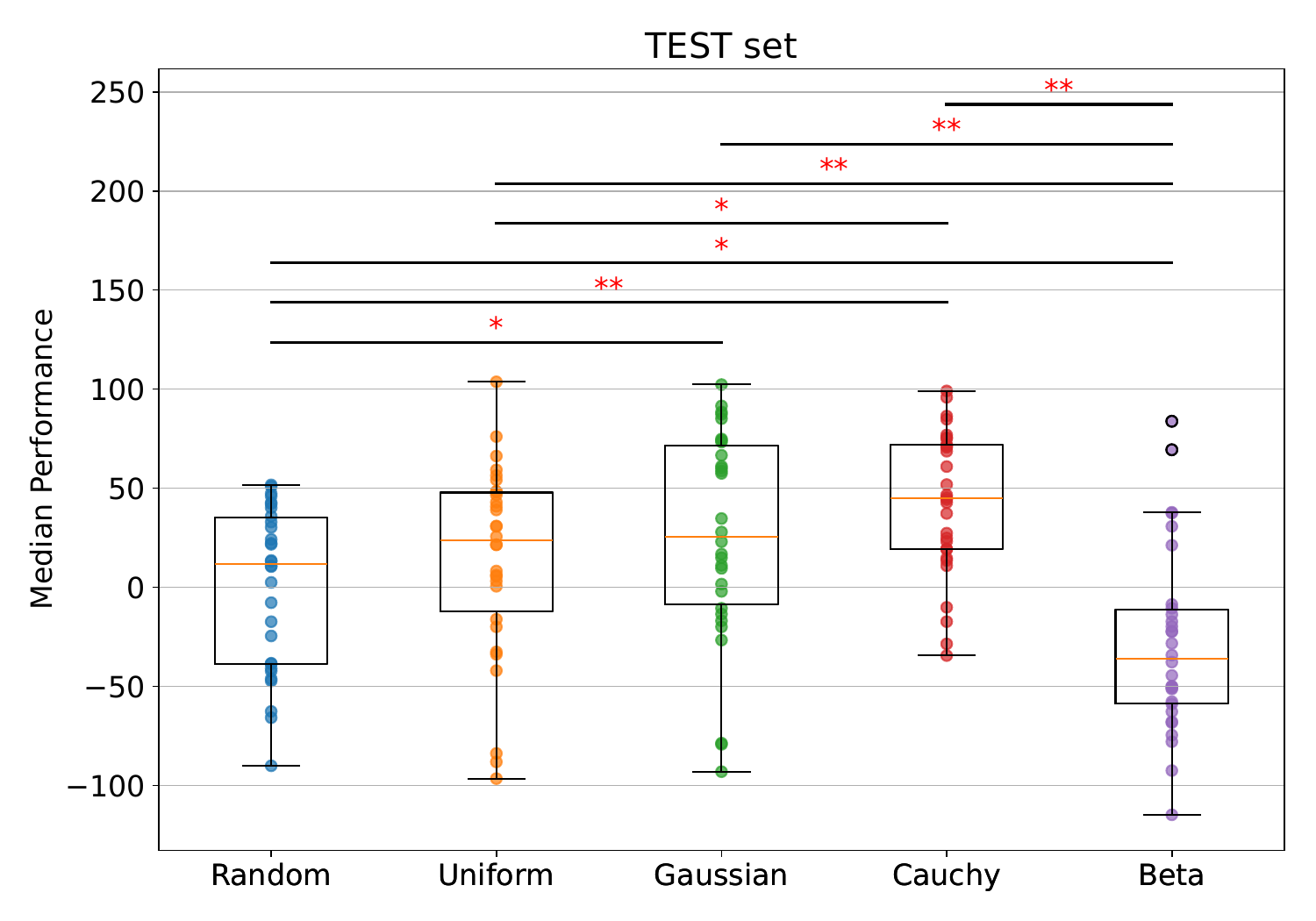}
 \caption{Performance comparison of training schedules on the testing morphologies on Bipedal Walker. * indicates $p < 0.05$, ** indicates $p < 0.01$, denoting statistical significance levels.}
 \Description{}
 \label{fig:Box_heterogeneity}
\end{figure}

For this analysis, we consider only the Bipedal Walker task. 
Figure \ref{fig:Box_heterogeneity} illustrates the performance results on Bipedal Walker of the different training schedules based on sampling distributions on the testing morphologies that are not introduced during training.

The figure shows the average, lower, and upper quartile obtained by $30$ independent evolutionary runs. The Beta training schedule obtains better performance on the testing morphologies (assuming minimization), with a median score of $-35.99$. The p-values show that there is a statistically significant difference between Beta and all the other distributions. 

Analyzing the performance of the morphologies on the training morphologies, we observe a different trend where the Uniform distribution stands out as the most effective, with a median performance of $-254.40$, and Beta as the worst one, with a median performance of $-218.66$. However, a statistically significant difference exists only between the Uniform and Beta distributions, evidenced by a p-value of $0.004$.

While the differences between these training schedules on the training morphologies are not as pronounced, we see that on the testing morphologies, the Beta training schedule leads to more generalization. This can be attributed to the fact that morphologies sampled with the Beta distribution lie closer to the ``border'' of the training parameter space, thereby resembling the morphologies encountered during testing. 

Similarly, if we compare the results on all morphologies combining the training and test sets, the Beta training schedule still demonstrates superior performance, achieving a score of $-123.55$, with a statistically significant difference from all other approaches. This aligns with our earlier discussion: introducing increased variability during training, especially in the bordering cases, tends to improve the controller's ability to extend its performance to unfamiliar scenarios. 

\subsection{Effect of Scheduling of Morphologies}
\label{sec:scheduling}
This section shows the results of the investigation on whether altering the order in which morphology variations are introduced during the evolutionary process yields different outcomes in terms of robustness and generalization. The central question we aim to address through this experiment is: ``Does modifying the order of samples impact learning and generalization?''
In order to answer this question, we conducted a comparative analysis between two training methodologies, namely the Random and the Incremental training schedules. In terms of heterogeneity, the training morphologies utilized are the same for both schedules. 
For this analysis, we consider Bipedal Walker and Walker2D.


\subsubsection*{Bipedal Walker}
\begin{figure}[ht!]
 \centering
 \includegraphics[clip, trim=0cm 0cm 0cm 1.2cm, width=\linewidth]{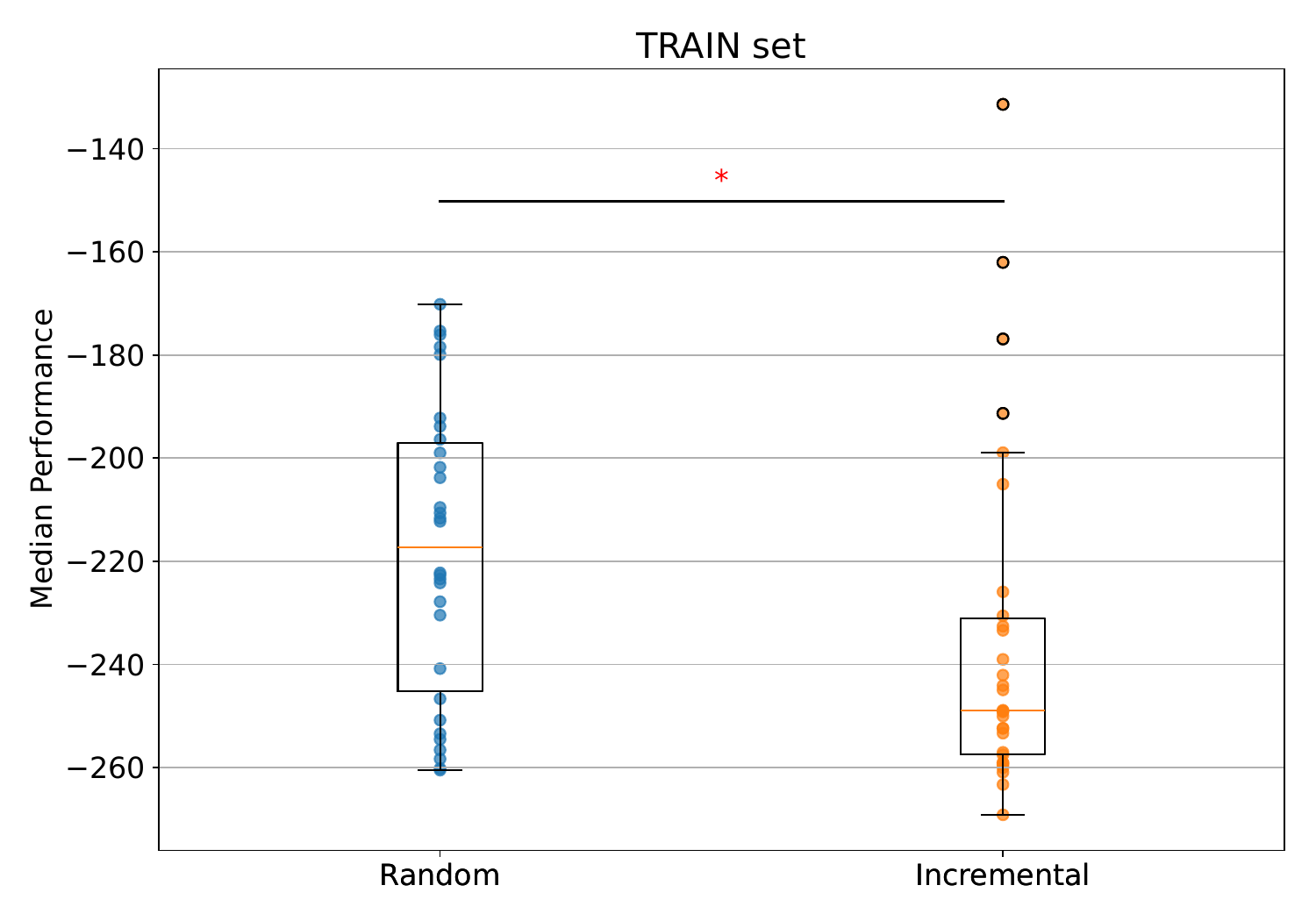}
 \caption{Performance comparison on the training morphologies on Bipedal Walker.  * indicates $p < 0.05$, ** indicates $p < 0.01$, denoting statistical significance levels.}
 \Description{}
 \label{fig:train_performance_schedule}
\end{figure}

On the Bipedal Walker, the controllers trained using the \textit{Incremental} training schedule achieved a median score of $-248.94$ on the training morphologies while scoring a median of $-2.21$ on the testing morphologies. In contrast, the controllers trained with the Random schedule achieved scores of $-217.25$ and $11.92$ on the training and testing morphologies, respectively. 
Although the performances on the testing morphologies are comparable, with no significant difference, a notable disparity emerges on the training morphologies, further evidenced by a $p$-value of $0.01$. Figure \ref{fig:train_performance_schedule} illustrates these results.
As can be seen from the figure, the controllers obtained with the Incremental training morphologies exhibit superior performance on the training morphologies compared to the one obtained with the Random schedule.

\subsubsection*{Walker2D}
We observe similar trends on the Walker2D task. Notably, the controllers trained with the Incremental training schedule led to a better performance compared to those trained with the Random schedule on both the training and testing morphologies. Specifically, they achieved a median score of $-1450.41$ and $-991.89$ on the training and testing morphologies, respectively, while the Random schedule yielded scores of $-1094.39$ and $-887.98$. However, no statistically significant difference is observed in both sets. 
\smallskip

These results highlight the significant impact of the order in which morphologies are presented during training. It is worth emphasizing once more that the set of morphologies remains identical; only the selection order differs. It is very interesting to observe the significant impact on performance that arises simply from altering the order in which morphologies are presented. Incrementally increasing the values of body parameters fosters the development of controllers that exhibit greater resilience to changes in morphologies, as also found in \cite{triebold2023evolving}.

\subsection{Multi-armed Bandit Approach}
\begin{figure*}[ht!]
 \includegraphics[width=\linewidth]{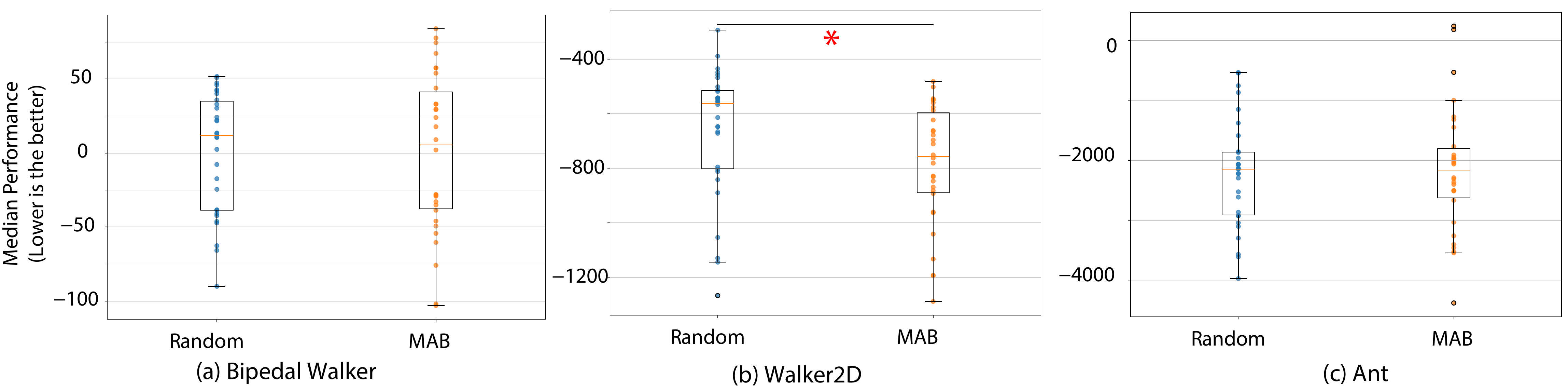}
 \caption{Performance comparison of Random schedule and the Multi-Armed Bandit approach on the testing morphologies on Bipedal Walker, Walker2D, and Ant.  * indicates $p < 0.05$, ** indicates $p < 0.01$, denoting statistical significance levels.}
 \Description{}
 \label{fig:MABcomparison}
\end{figure*}

This section presents the results of the MAB approach for morphology selection during training, taking the Random schedule as the baseline. It is important to highlight that all the training morphologies are the same; the only difference lies in the order in which morphologies are selected. In this set of experiments, we consider the three tasks, namely Bipedal Walker, Walker2D, and Ant.
Figure \ref{fig:MABcomparison} visualizes these results.



\paragraph{Bipedal Walker}
Although the difference was not statistically significant, the MAB approach achieved a better median score of $-245.41$ on training (relative to $-217.26$ for Random) 
and a median score of $5.52$ on testing morphologies (relative to $11.92$ for Random) 
On the training morphologies, the ANN controllers trained with the MAB approach demonstrated performance superior to the Random schedule, meaning that the MAB strategy used for training was effective. 
On the testing morphologies instead, it reached comparable performances with Random.

\paragraph{Walker2D}
In the Walker2D task, the MAB approach was significantly better relative to Random. It achieved a median score of $-1433.21$ on the training morphologies, compared to $-1325.01$ for the Random schedule. 
Conversely, on the testing morphologies, it reached a median score of $-756.76$, whereas the Random schedule scored $-562.05$. 
Remarkably, on both the training and testing morphologies, the network trained with the MAB approach outperformed
the Random training schedule, with a statistically significant difference on the test set. 

\paragraph{Ant}
Finally, on the Ant task, the MAB approach showed superior performance when compared to the Random schedule, both in the training and testing morphologies, although statistical significance was not achieved. On the training morphologies, the MAB approach achieved a median score of $-3824.95$, compared to $-3779.36$ obtained by the Random schedule. On the testing morphologies, instead, the MAB approach yields a median score of $-2169.75$, while the Random schedule achieves $-2139.04$. 

\begin{figure}[!ht]
 \hspace{0.5cm} 
 \centering
 \includegraphics[width=\linewidth]{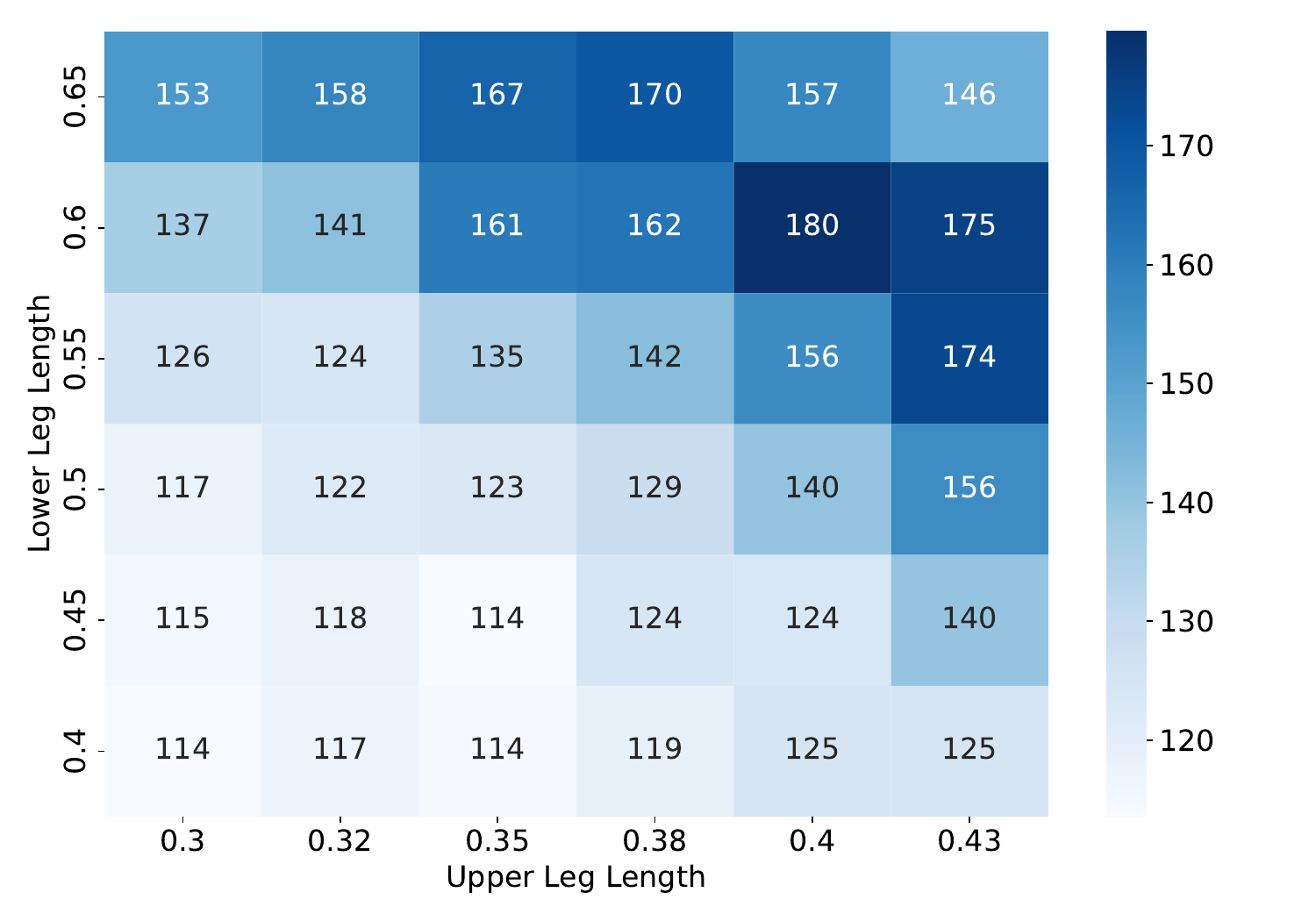}
 \caption{Heatmap illustrating the frequency of selection for each morphology during training on Walker2D. Each cell in the heatmap represents a unique morphology, and the numerical value within the cell indicates the frequency with which that morphology was selected during training. This frequency is computed as the average occurrence across 30 independent training runs.}
 \Description{}
 \label{fig:env_used_walker}
\end{figure}

\subsection{Analysis on Morphology Selection}
An interesting analysis to perform is to look at which morphologies are selected most frequently, averaged over the $30$ runs, during training. 

\begin{figure*}[ht!]
 \centering
 \begin{subfigure}{.45\textwidth}
 \centering
 \includegraphics[width=\linewidth]{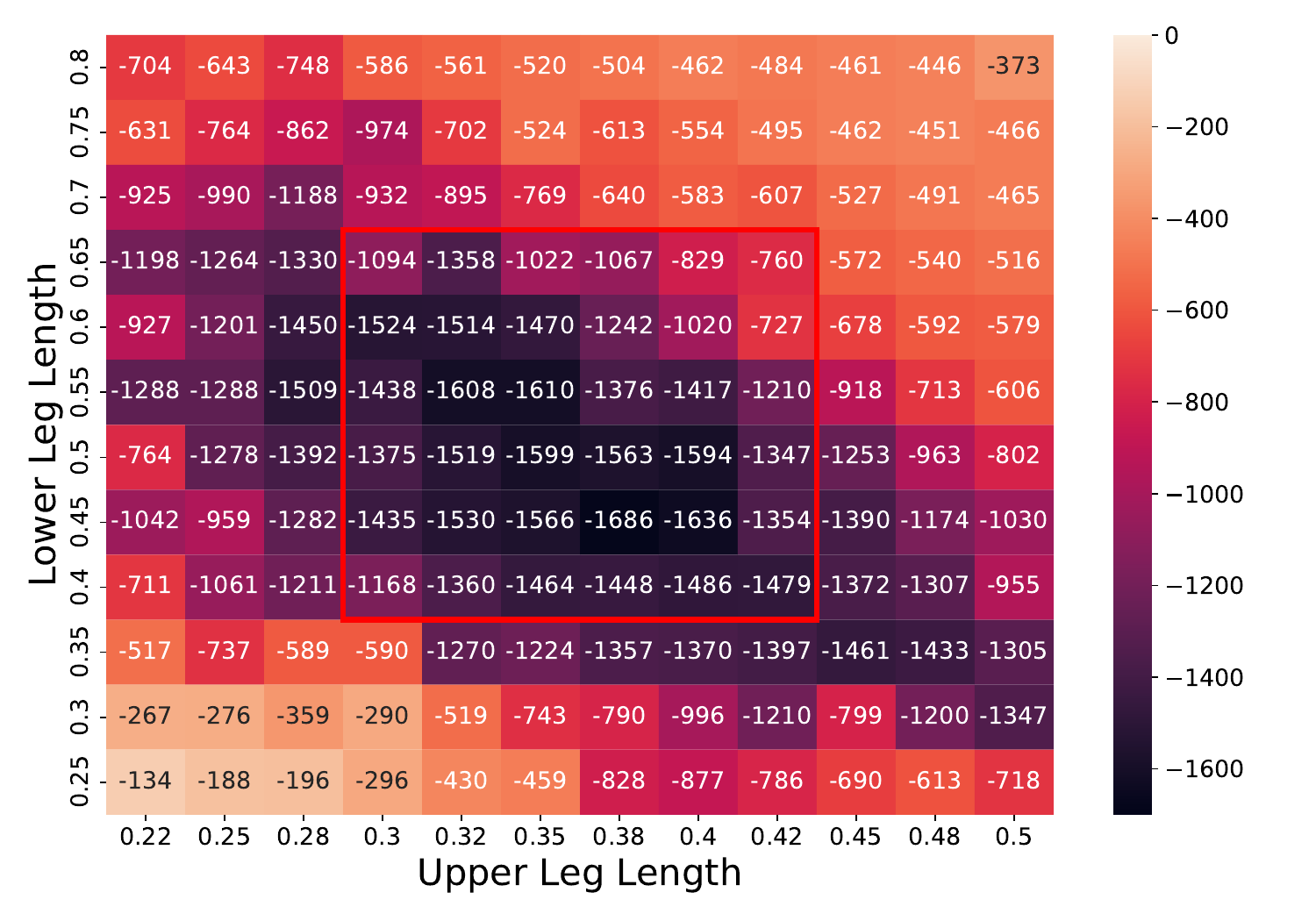}
 \caption{Random Schedule}
 \Description{}
 \label{fig:MAB_Walker_random}
 \end{subfigure}%
 \begin{subfigure}{.45\textwidth}
 \centering
 \includegraphics[width=\linewidth]{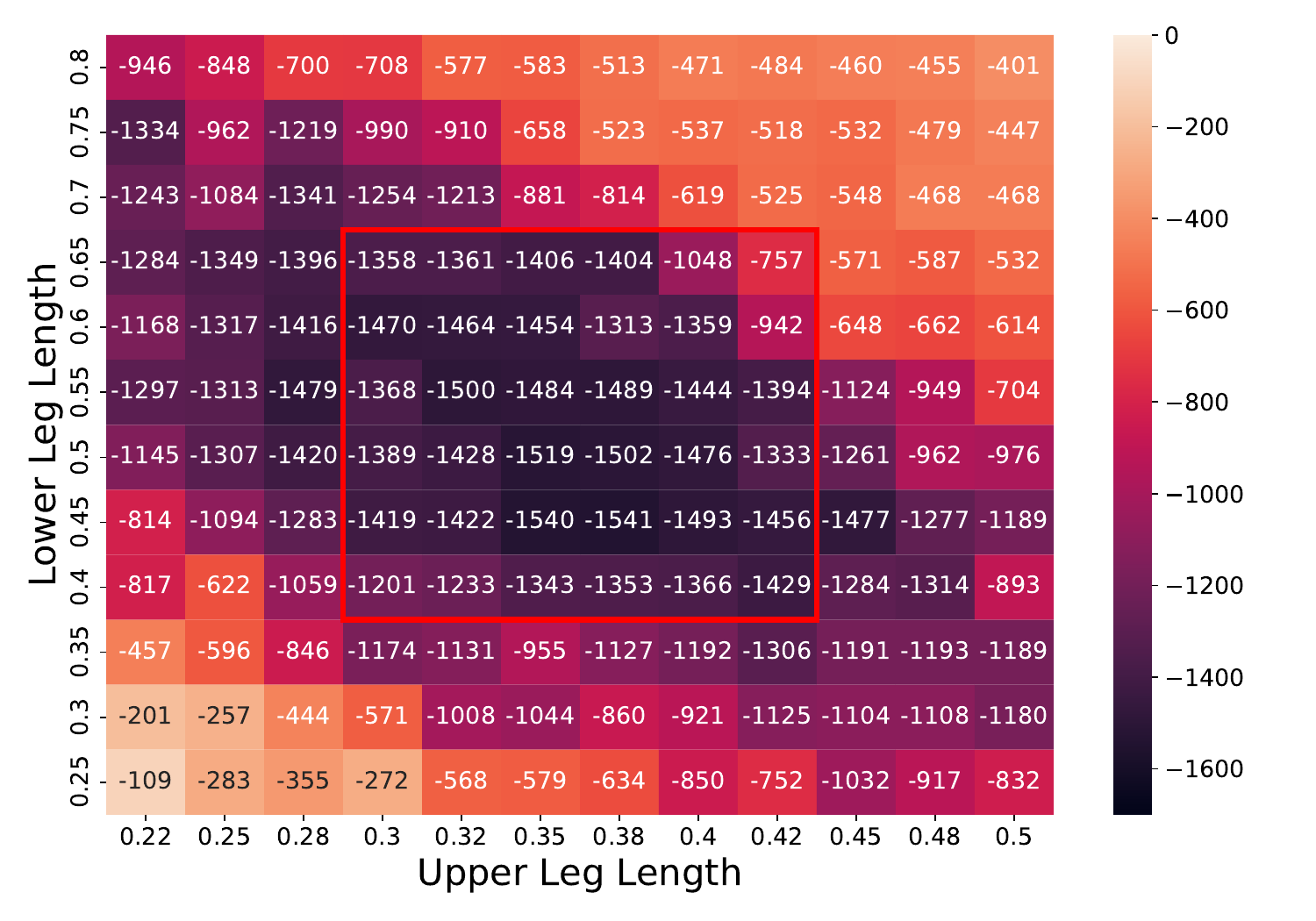}
 \caption{Multi-Armed Bandit Approach}
 \Description{}
 \label{fig:MAB_Walker_MAB}
 \end{subfigure}
 \caption{Heatmap illustrating the average performance achieved on all morphological variations on Walker2D using the Random and MAB training schedules. The cells that are within the square with red borders show the training morphologies. The morphological parameters that are beyond these borders are used for the testing morphologies.}
 \label{fig:heatmap_MAB36_walker}
\end{figure*}

\paragraph{Walker2D}
Figure \ref{fig:env_used_walker} presents a heatmap illustrating the frequency of each morphology's usage for Walker2D. The morphologies selected the most are represented by darker colors. Analysing the heatmap we can see that the morphologies selected the most are the ones in the upper-right region of the morphological parameter space. 
The MAB approach exhibits a non-uniform selection pattern for morphologies, particularly focusing on the upper-right quadrant (morphologies with high upper leg length and high lower leg length). The morphology selected the most is the one with an upper leg length of $0.4$ and lower leg length of $0.6$, used $180$ times, while the least used was used $114$ times. The MAB approach identifies this region as prone to improvements, while other regions (such as the lower-left quadrant) may present challenges, resulting in limited enhancements.

Figure \ref{fig:heatmap_MAB36_walker} presents heatmaps illustrating the average performances across all morphologies obtained using two different scheduling strategies: the Random schedule (Figure \ref{fig:MAB_Walker_random}) and the MAB approach (Figure \ref{fig:MAB_Walker_MAB}). The same analyses for Bipedal Walker and Ant are reported in the Supplementary Material. 
Each cell in the heatmaps corresponds to a specific morphology. Unlike the frequency heatmap, this visualization incorporates morphologies from both the training (inside the red square) and testing morphologies. The shading of each cell indicates the average performance of the ANN controllers on that particular morphology, with darker shades indicating better scores. These averages are computed over all controllers obtained from $30$ runs. 
Observing the performance on the training morphologies, in the case of the MAB approach, the most significant improvement is notably concentrated in the upper-right region of the space (within the red square), coinciding with the most frequently chosen morphologies. This indicates that the MAB approach has effectively enhanced performance in a specific area of the space where improvement was feasible. Selecting more morphologies in that region of the space allowed to evolve ANN-based controllers with a greater capability to control those types of morphologies, without affecting performances in other parts of the morphological space.


\paragraph{Bipedal Walker}
In Bipedal Walker (heatmap reported in the Supplementary Material), the MAB approach favors morphologies located in the lower-right region of the morphology space (high leg width and low leg length). Morphologies in this region are utilized roughly twice as often as those in the left upper quadrant. For instance, the morphology with leg width of $17$ and leg height of $24$ was selected an average of $135$ times, whereas the morphology with leg width of $7$ and leg height of $44$ was chosen an average of $61$ times. While it may seem logical to assume that the lower-right part was prioritized because it is the more complex part of the morphology space to control, this interpretation is not entirely accurate. The most challenging area within the morphology space is not the lower-right part, but rather the upper-right section appears to pose greater difficulty for the ANN-based controller. This is also evident from the heatmaps provided in the Supplementary Material, where the controller achieves higher scores in the upper right region, indicating poorer performance. The focus on training with morphologies in the lower right section could be attributed to the potential for performance improvement in this area. In contrast, the upper right section may have reached a point where further improvements are no longer possible. Therefore it may be beneficial to focus on that part because improvements are more readily achievable. Overall, on this task the strategy employed by the MAB approach improves the performances compared to the Random training schedule, focusing on a region of the morphology space in which performances can be increased without negatively affecting performances in other parts of the space.

\paragraph{Ant}
Upon analyzing the frequency of morphology selection during training on Ant, we observed an imbalance in the utilization of morphologies also in this case, particularly skewed towards the left-central region of morphology space (heatmap reported in the Supplementary Material). The focus of the MAB approach on this specific region of the space leads to an improvement in performance within that area. The controllers trained with the Random schedule achieved higher scores predominantly on the right side of the morphology space, whereas the networks trained with the MAB schedule slightly enhanced performance towards the left region. Nonetheless, this improvement comes at the expense of slightly degraded performance on the right side of the space. While we may not observe a significant difference in results, it is interesting to observe that even in this case, the MAB approach chooses to focus on a specific region of the morphology space rather than uniformly utilizing all available morphologies.


\section{Conclusions}
\label{sec:concl}

In this study, we examined the impact of variability on the learning and generalization of ANN-based controllers. Alongside several predefined training schedules to select morphologies to train on, we employed a reinforcement learning-based approach to the automatic selection of these morphologies formalizing the selection as a multi-armed bandit problem. Our analysis focused on two types of variability: heterogeneity of samples and scheduling of samples. The analysis of different training schedules highlighted the importance of increasing heterogeneity for improved robustness and generalization. Notably, using a uniform distribution for training morphologies generation led to better performance on the training morphologies compared to using discrete morphologies. The Beta training schedule showed the best approach to enhance generalization to previously unseen morphologies. Furthermore, we explored how altering the scheduling of the training morphologies affects the results. In this case, the Incremental training schedule showed the best performance. The MAB approach for automatic morphology selection during training revealed a non-uniform selection strategy by focusing on certain regions, which in turn led to significant improvements relative to Random selection.

Future research can deepen our understanding of how variability influences generalization. Investigating discriminative and non-discriminative features within tasks may provide valuable insights; for instance, we found that modifying the leg length in Bipedal Walker had a greater impact on performance than the leg width, which suggests that this parameter is more discriminative. One approach could be extended to prioritize variability introduction on non-discriminative features before moving to discriminative ones, a strategy that has been shown to enhance learning \cite{RAVIV2022462}. Exploring automated scheduling through meta-learning approaches like (Deep) Q-learning is another avenue for investigation. Additionally, evolutionary algorithms hold the potential for optimizing the order of samples, though computational challenges must be addressed. Furthermore, future studies could compare the performance of non-stationary MAB approaches against stationary ones, shedding light on their relative effectiveness. These research directions offer promising avenues for advancing our understanding of variability and optimization in complex tasks.


\balance
\bibliographystyle{ACM-Reference-Format}
\bibliography{biblio.bib}

\appendix

\end{document}


\title{The Effect of Training Schedules in Morphological Generalization\\(Supplementary Material)}

\author{Edoardo Barba}
\affiliation{%
  \institution{University of Trento}
  \streetaddress{Via Calepina, 14}
  \city{Trento}
  \country{Italy}
  \postcode{38122}
}

\author{Anil Yaman}
\affiliation{%
  \institution{Vrije Universiteit Amsterdam}
  \streetaddress{De Boelelaan 1105}
  \city{Amsterdam}
  \country{Netherlands}
}

\author{Giovanni Iacca}
\affiliation{%
  \institution{University of Trento}
  \streetaddress{Via Calepina, 14}
  \city{Trento}
  \country{Italy}
  \postcode{38122}
}

\maketitle
\pagestyle{plain} 

\appendix

\section{Dataset visualization}
\subsection{Training and testing sets}
\begin{figure}[ht!]
    \centering
    \includegraphics[width=.7\linewidth]{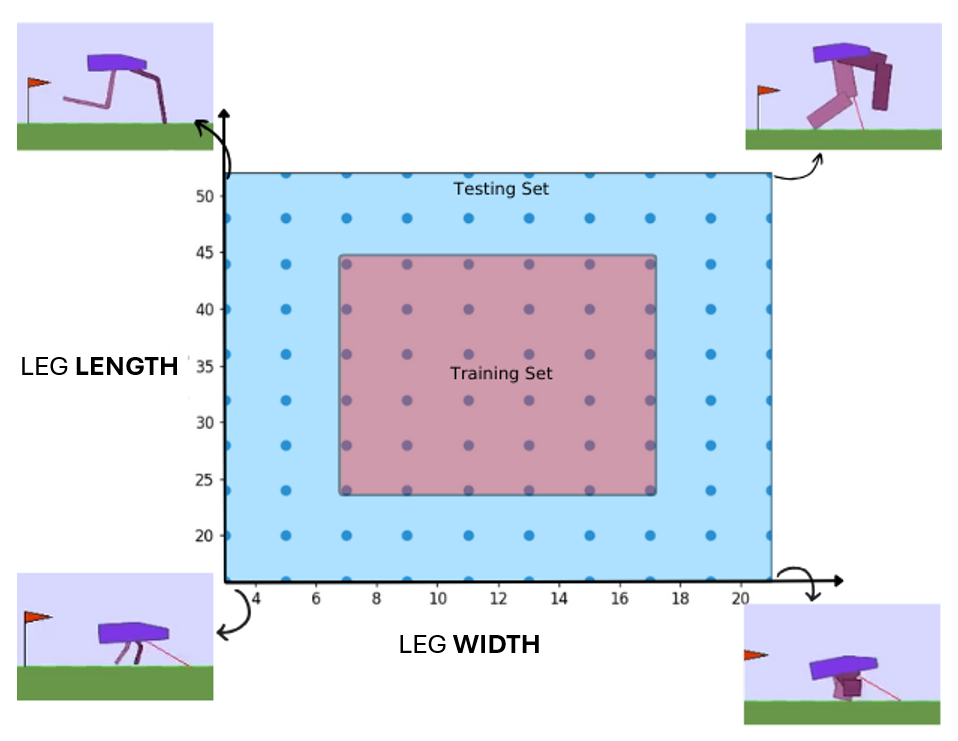}
    \caption{Visual representation of the Training and Testing sets for the Bipedal Walker task. Each point represents a specific morphology characterized by a combination of leg width and leg length parameters. The training set consists of 36 morphologies, while the testing set consists of 64 morphologies.}
    \Description{}
    \label{fig:train_test}
\end{figure}

Figure \ref{fig:train_test} presents a visual depiction of the training and testing sets for the Bipedal Walker task. The dots representing the training set are depicted with a red background, while those representing the testing set have a blue background. Each dot on the plot represents a specific morphology characterized by a combination of leg width and leg length parameters.

To assess the performance of an individual on the training set, its score is computed for each morphology within the training set. These individual scores are then averaged to derive a single metric representing the individual's performance on the training set. The same methodology is applied to compute the individual's score on the testing set. This systematic approach facilitates a thorough evaluation of the individual's ability to generalize across a diverse range of morphologies present in both the training and testing sets.

\subsection{Validation sets}

\begin{figure}[ht!]
    \centering
    \begin{subfigure}{0.33\textwidth}
    \centering
    \includegraphics[width=\linewidth]{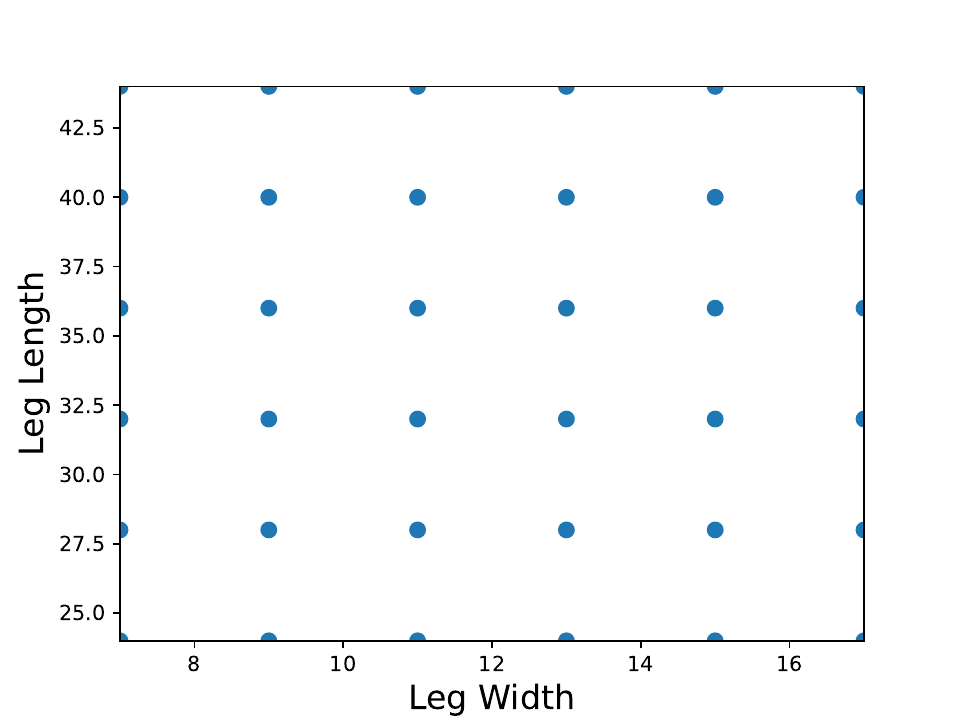}
    \caption{Bipedal Walker}
    \Description{}
    \label{fig:val_set_biped}
    \end{subfigure}
    \hfill
    \begin{subfigure}{0.33\textwidth}
    \centering
    \includegraphics[width=\linewidth]{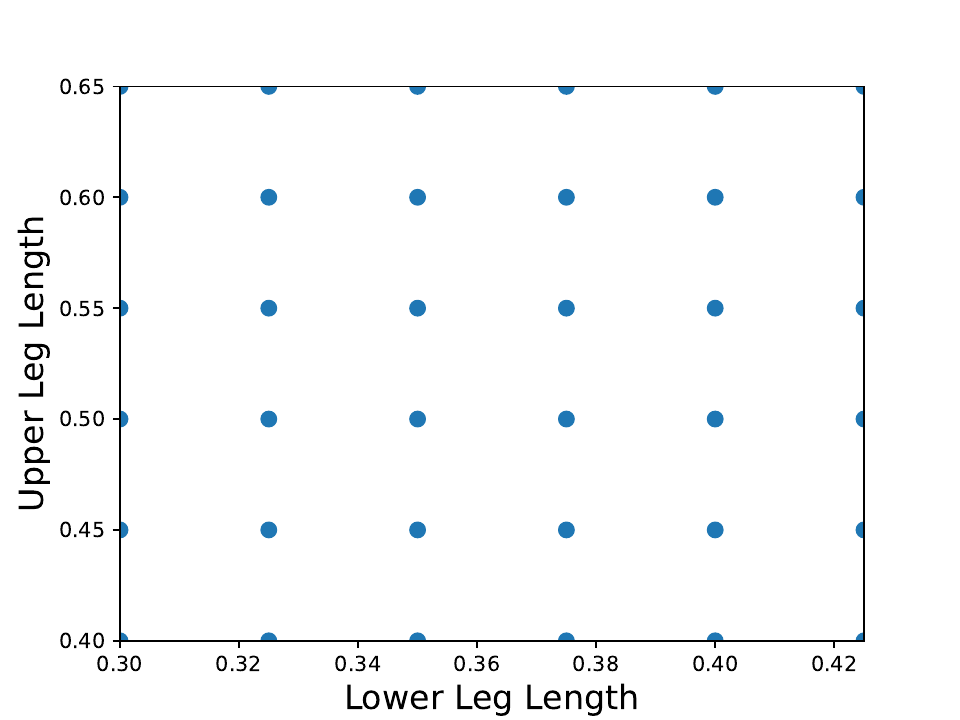}
    \caption{Walker2D}
    \Description{}
    \label{fig:val_set_walker}
    \end{subfigure}
    \hfill
    \begin{subfigure}{0.33\textwidth}
    \centering
    \includegraphics[width=\linewidth]{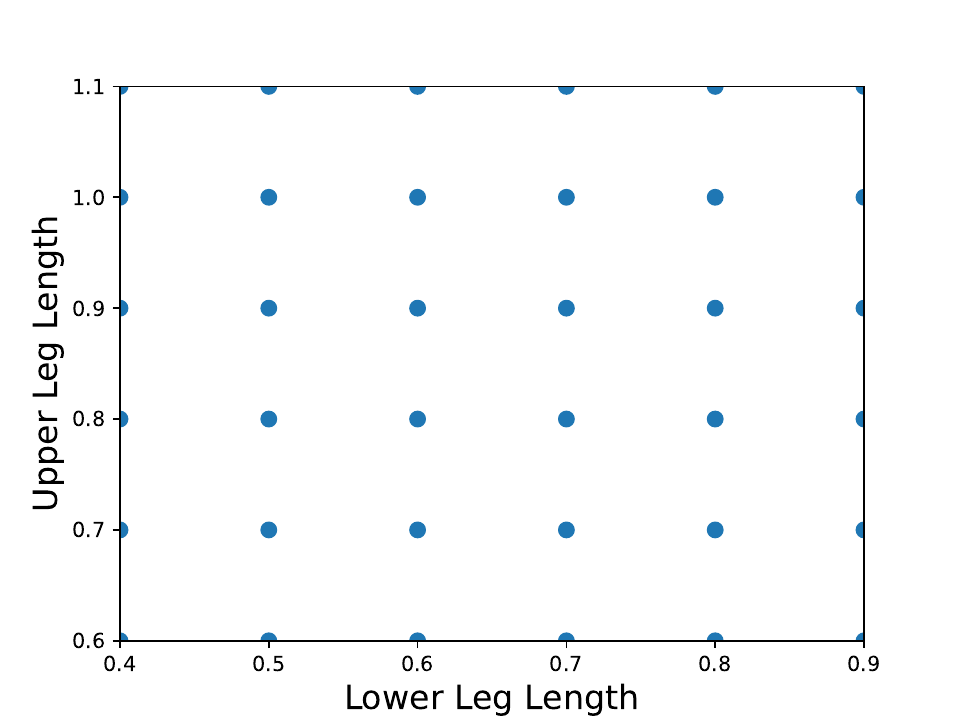}
    \caption{Ant}
    \Description{}
    \label{fig:val_set_ant}
    \end{subfigure}
    \caption{Validation sets on the three tasks.}
\end{figure}

Figures \ref{fig:val_set_biped}-\ref{fig:val_set_ant} illustrate the validation sets utilized during the training phase for the respective tasks. Each point in these plots represents a morphology characterized by specific parameter combinations. During training, the validation set serves to evaluate the generalization ability of the top-performing individual on the morphology presented in that generation. 
This process entails computing the score achieved by the individual on each of these 36 morphologies separately and then averaging these scores to obtain a single metric.

\section{Distribution visualization}
\begin{figure}[ht!]
    \centering
    \begin{subfigure}{0.45\textwidth}
    \centering
    \includegraphics[width=\linewidth]{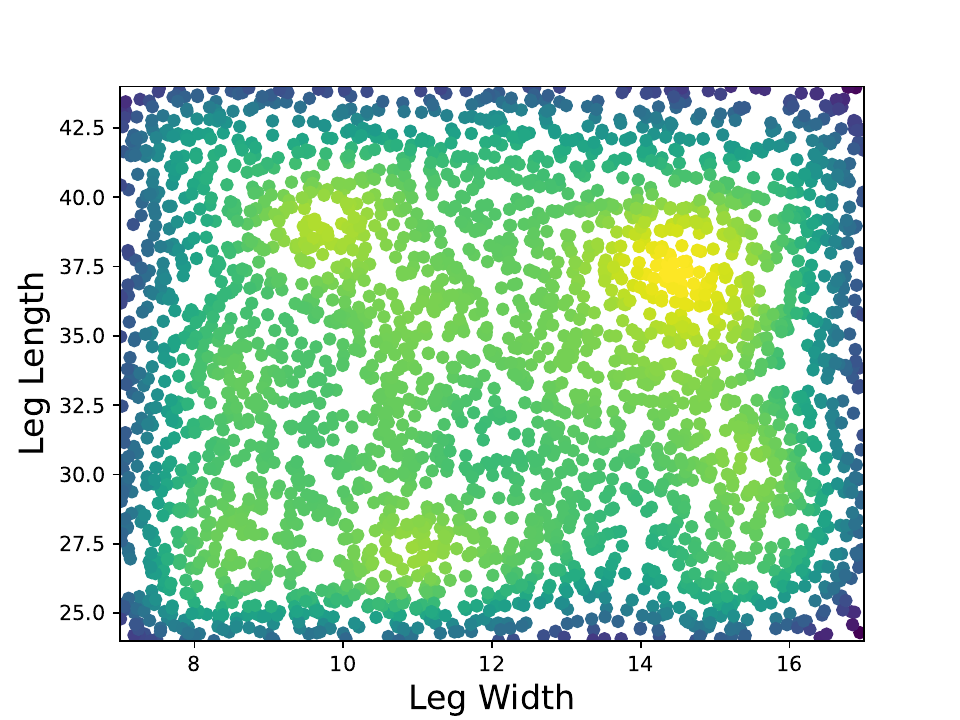}
    \caption{Uniform distribution}
    \Description{}
    \label{fig:unif_distrib}
    \end{subfigure}
    \hfill
    \begin{subfigure}{0.45\textwidth}
    \centering
    \includegraphics[width=\linewidth]{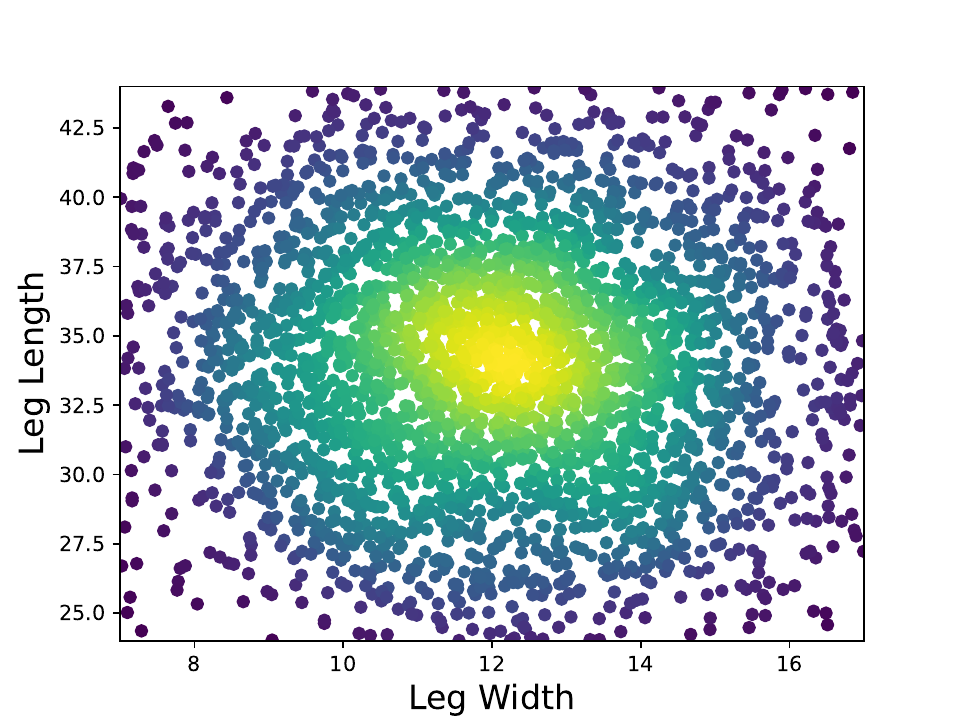}
    \caption{Gaussian distribution}
    \Description{}
    \label{fig:gauss_distrib}
    \end{subfigure}
    
    \begin{subfigure}{0.45\textwidth}
    \centering
    \includegraphics[width=\linewidth]{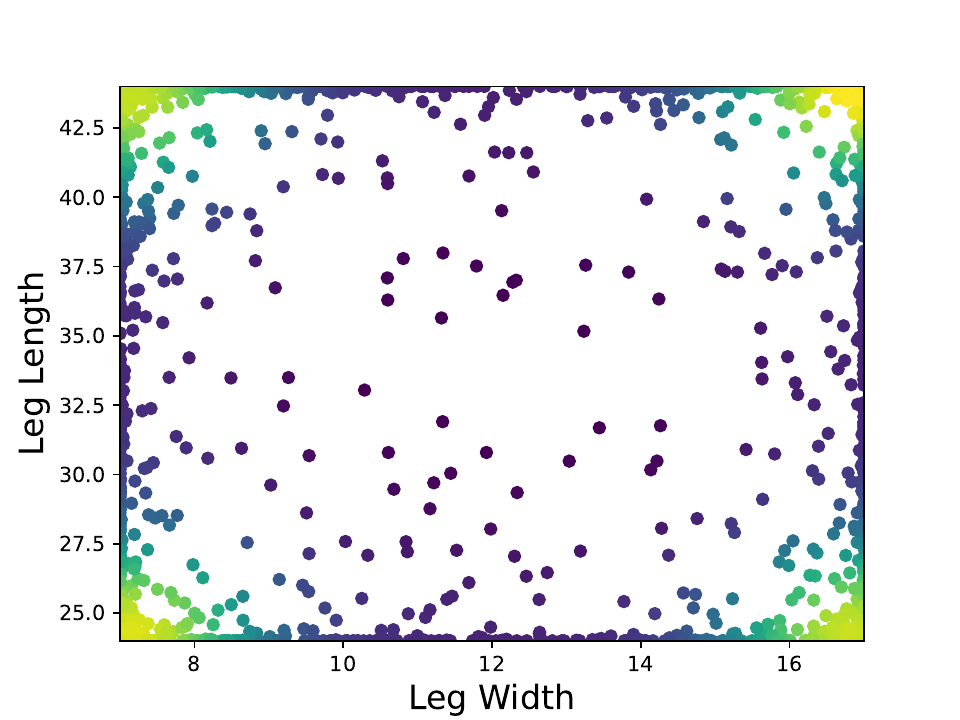}
    \caption{Beta distribution}
    \Description{}
    \label{fig:beta_distrib}
    \end{subfigure}
    \hfill
    \begin{subfigure}{0.45\textwidth}
    \centering
    \includegraphics[width=\linewidth]{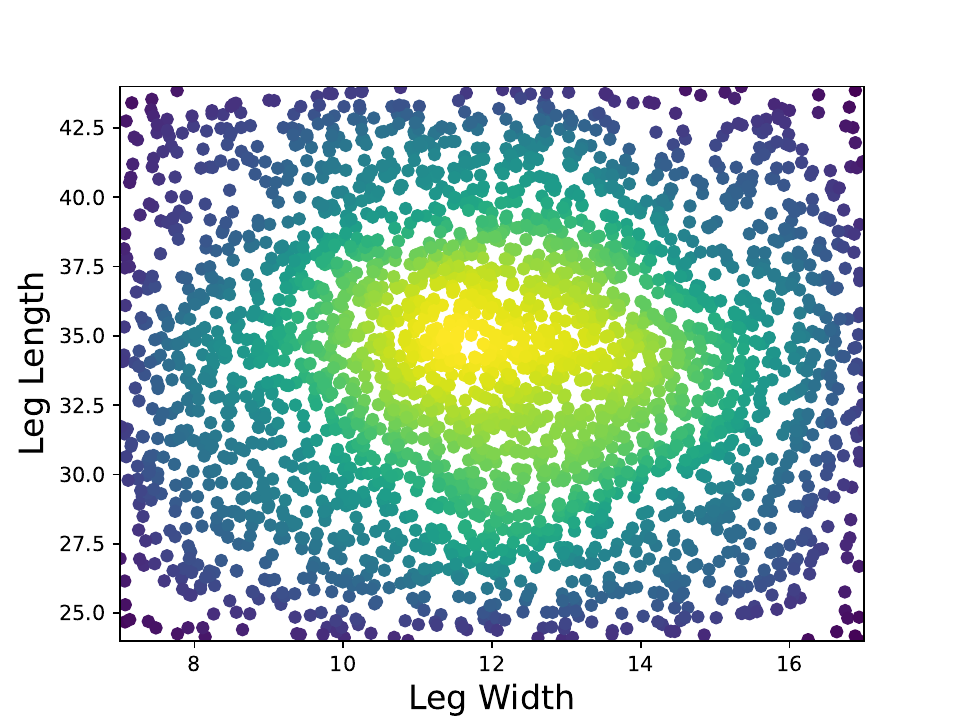}
    \caption{Cauchy distribution}
    \Description{}
    \label{fig:cauchy_distrib}
    \end{subfigure}
    \caption{Distribution on Ant.}
\end{figure}

Figures \ref{fig:unif_distrib}-\ref{fig:cauchy_distrib} provide visual representations of the training sets utilized on the Ant task by the Uniform, Gaussian, Beta, and Cauchy training schedules, respectively. In these plots, each point signifies a morphology, with parameters sampled independently from the given distribution. The background colors serve to reflect the density of the samples, providing insights into the distribution of morphologies across the parameter space.

\section{Performance analysis}
\begin{figure}[ht!]
    \centering
    \begin{subfigure}{0.45\textwidth}
    \centering
    \includegraphics[width=\linewidth,clip,trim=0 0 0 1.2cm]{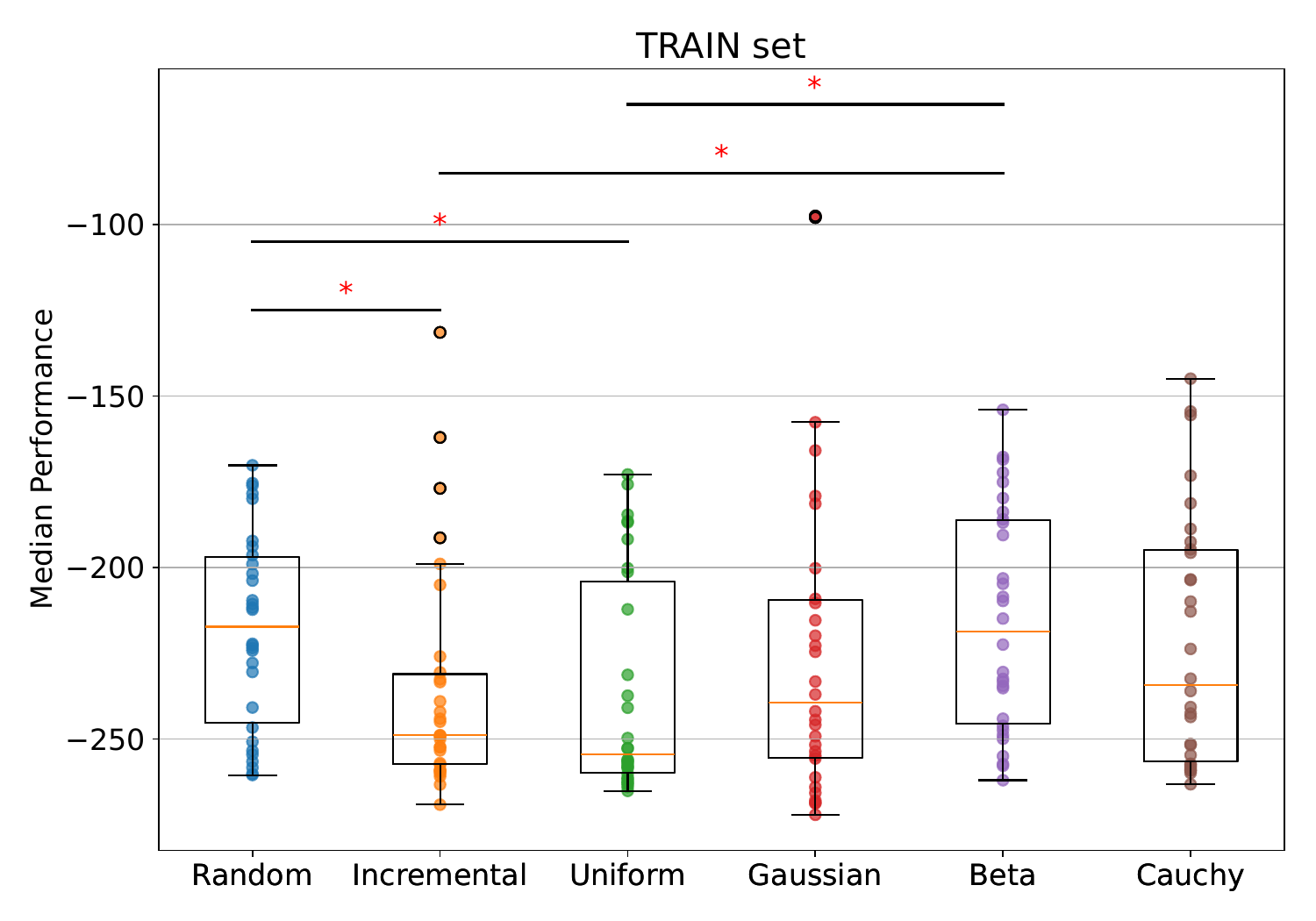}
    \caption{Training set}
    \Description{}
    \label{fig:train_performance_all}
    \end{subfigure}
    \hfill
    \begin{subfigure}{0.45\textwidth}
    \centering
    \includegraphics[width=\linewidth,clip,trim=0 0 0 1.2cm]{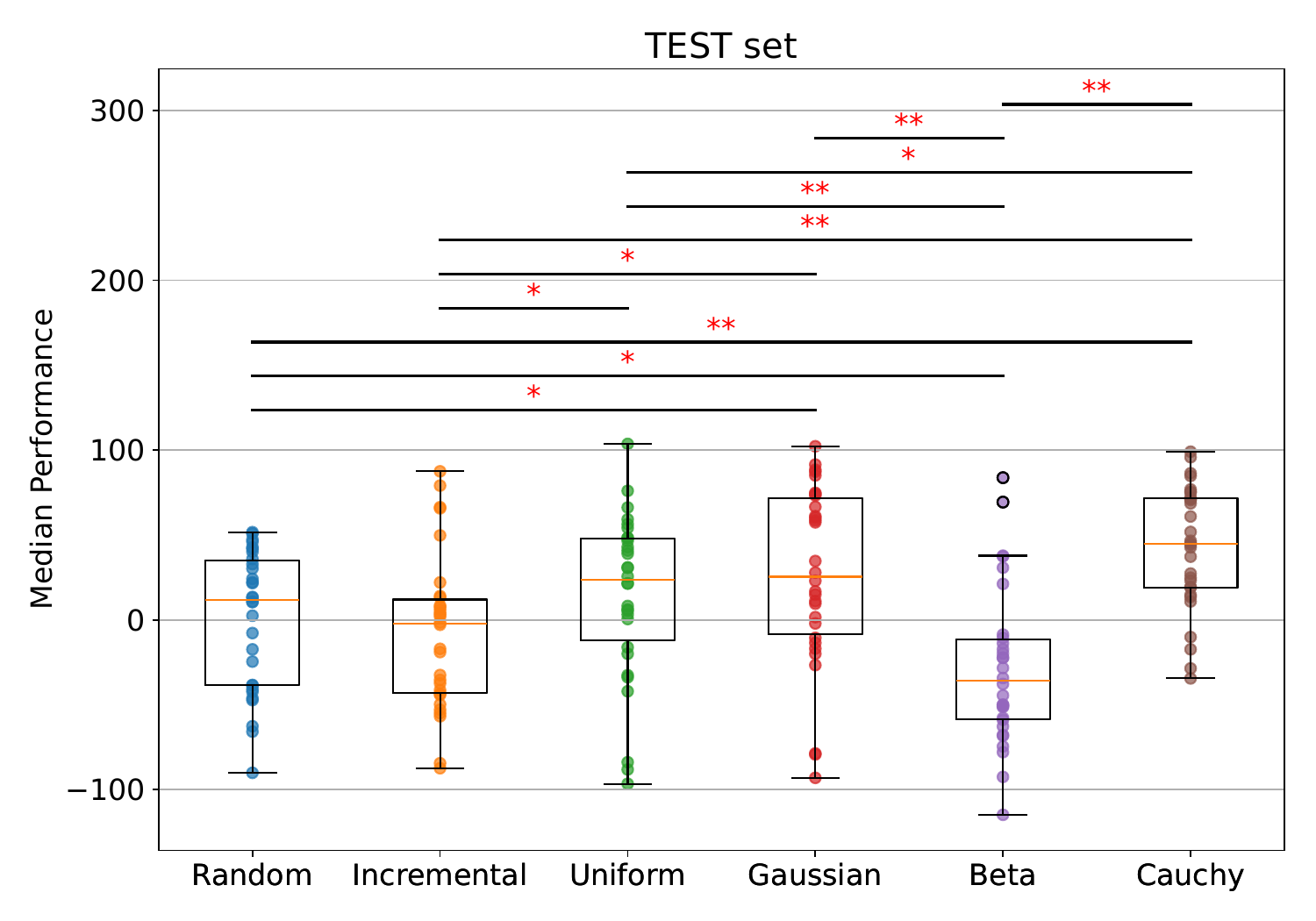}
    \caption{Testing set}
    \label{fig:test_performance_all}
    \Description{}
    \end{subfigure}
    \caption{Performance of all the training schedules on Bipedal Walker. * and ** indicate statistical significance with $p < 0.05$ and $p < 0.01$, respectively.}
\end{figure}

Figures \ref{fig:train_performance_all} and \ref{fig:test_performance_all} present boxplots representing the performance of artificial neural networks (ANNs) trained using various training schedules on Bipedal Walker, respectively on the training and testing sets.

\subsection{Scheduling}
\begin{figure}[ht!]
    \centering
    \begin{subfigure}{0.45\textwidth}
    \centering
    \includegraphics[width=\linewidth,clip,trim=0 0 0 1.2cm]{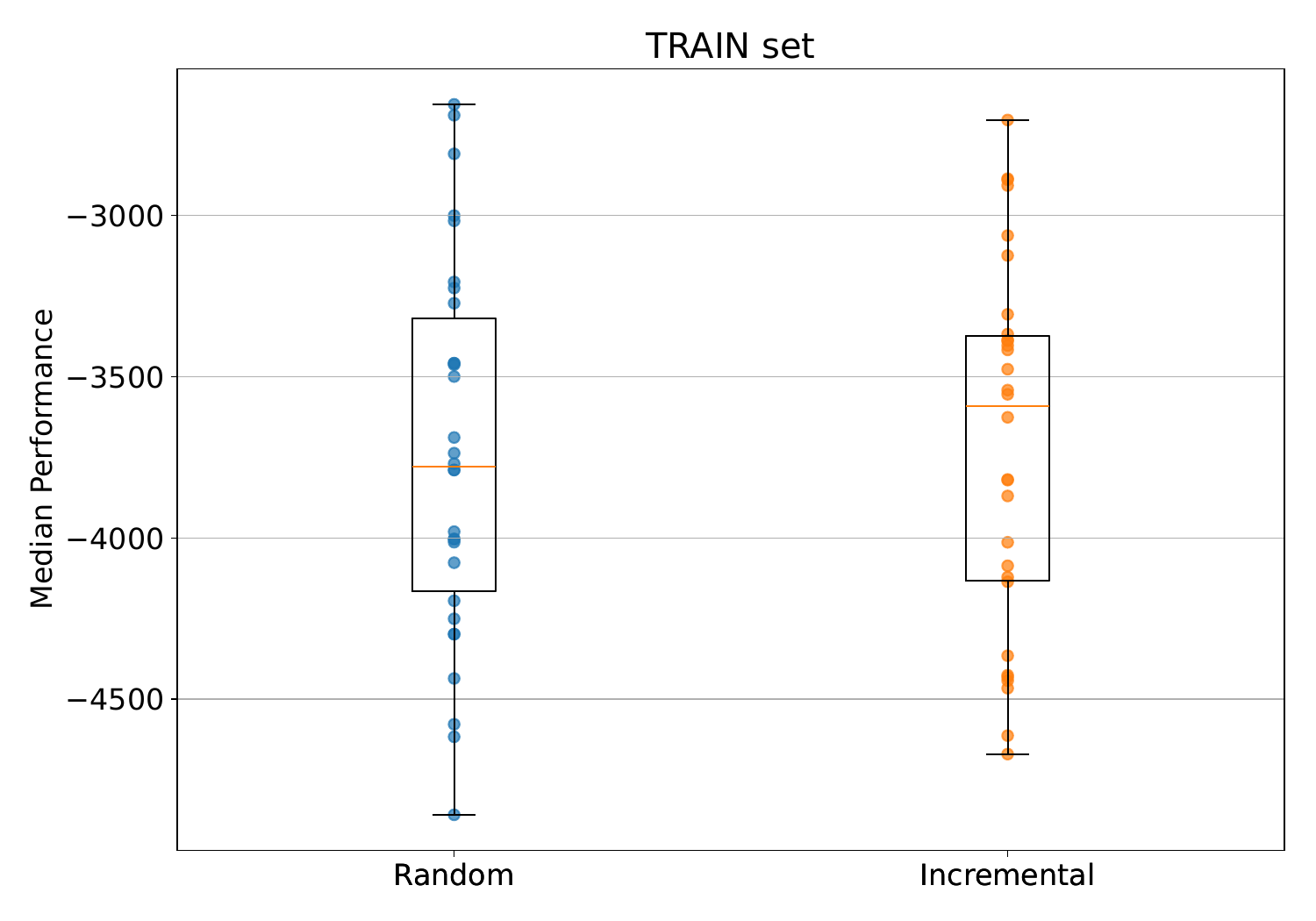}
    \caption{Training set}
    \Description{}
    \label{fig:train_performance_ant_scheduling}
    \end{subfigure}
    \hfill
    \begin{subfigure}{0.45\textwidth}
    \centering
    \includegraphics[width=\linewidth,clip,trim=0 0 0 1.2cm]{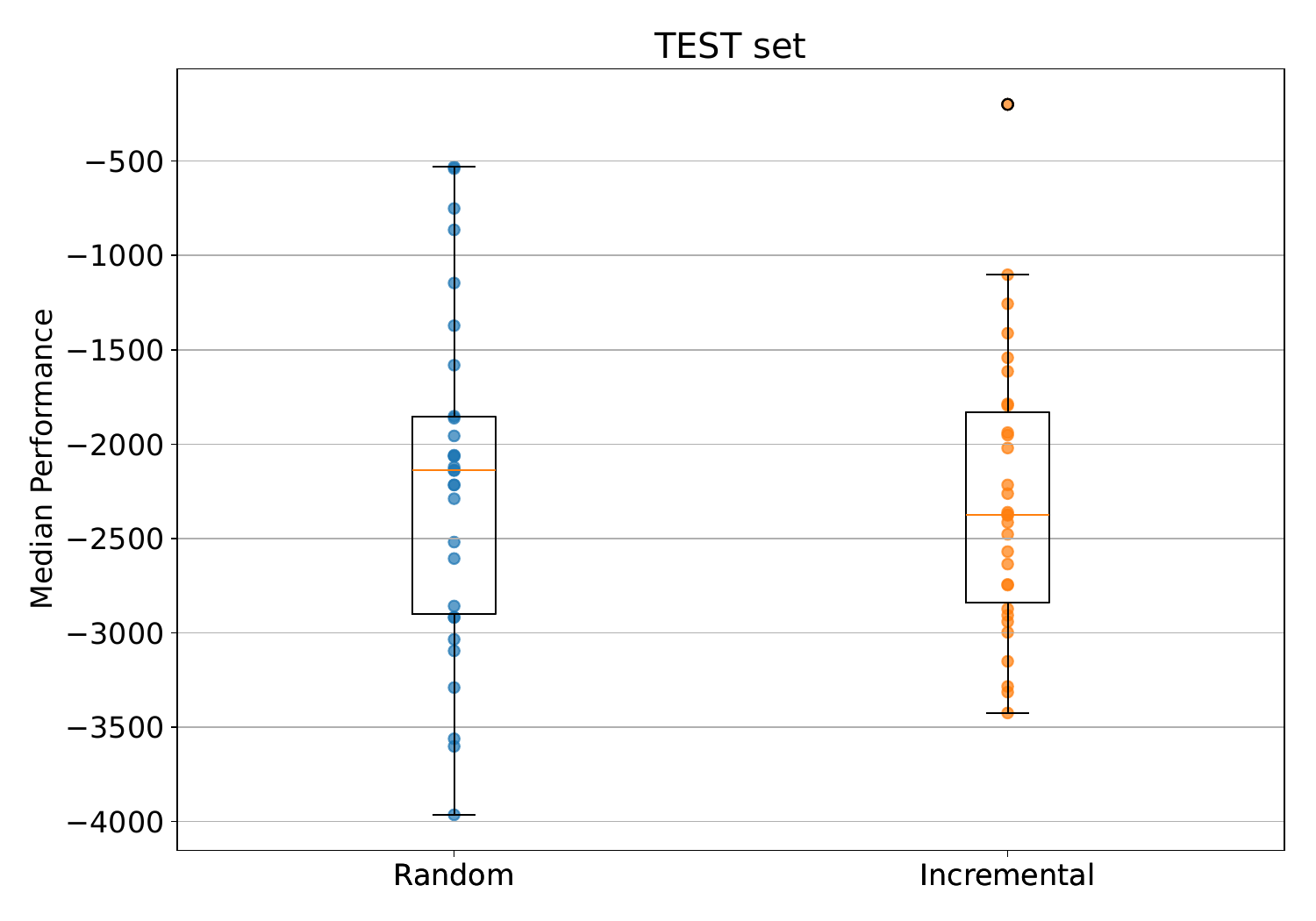}
    \caption{Testing set}
    \Description{}
    \label{fig:test_performance_ant_scheduling}
    \end{subfigure}
    \caption{Performance comparison between Random and Incremental schedules on Ant.}
\end{figure}

Figures \ref{fig:train_performance_ant_scheduling} and \ref{fig:test_performance_ant_scheduling} show the performance of ANNs trained using Random schedule and Incremental schedule on the Ant task, respectively on the training and testing sets. The network trained with the Random schedule achieved a median performance of -3779.36 on the training set and -2139.04 on the testing set. Conversely, the network trained with the Incremental schedule reached -3590.29 and -2375.35 on the respective sets.

\section{Multi-Armed Bandit schedule}

\subsection{Performance comparison}

\begin{figure}[ht!]
 \centering
 \begin{subfigure}{.45\textwidth}
 \centering
 \includegraphics[width=\linewidth,clip,trim=0 0 0 1.2cm]{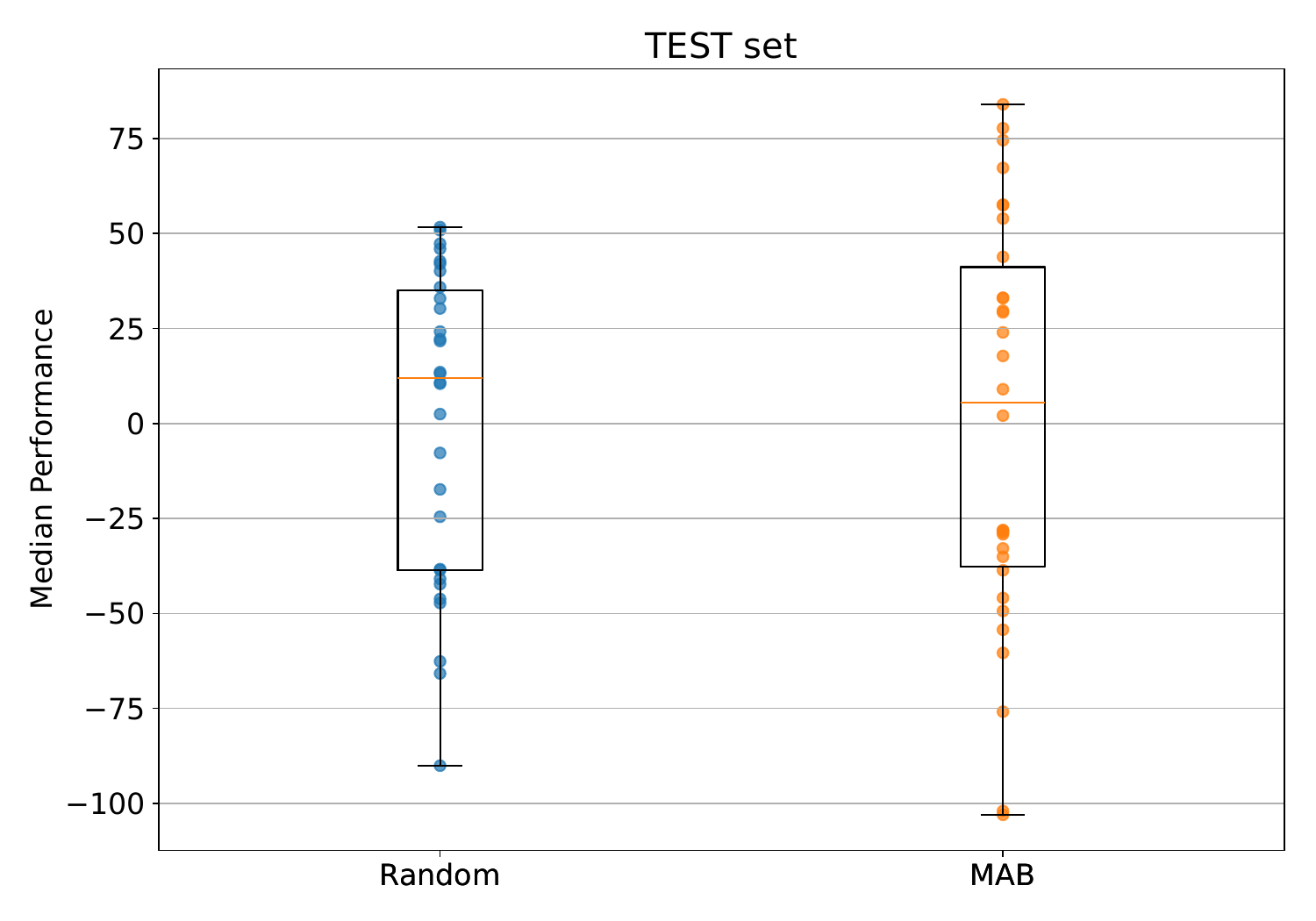}
 \caption{Bipedal Walker}
 \Description{}
 \label{fig:Box_MAB_biped}
 \end{subfigure}
 \hfill
 \begin{subfigure}{.45\textwidth}
 \includegraphics[width=\linewidth,clip,trim=0 0 0 1.2cm]{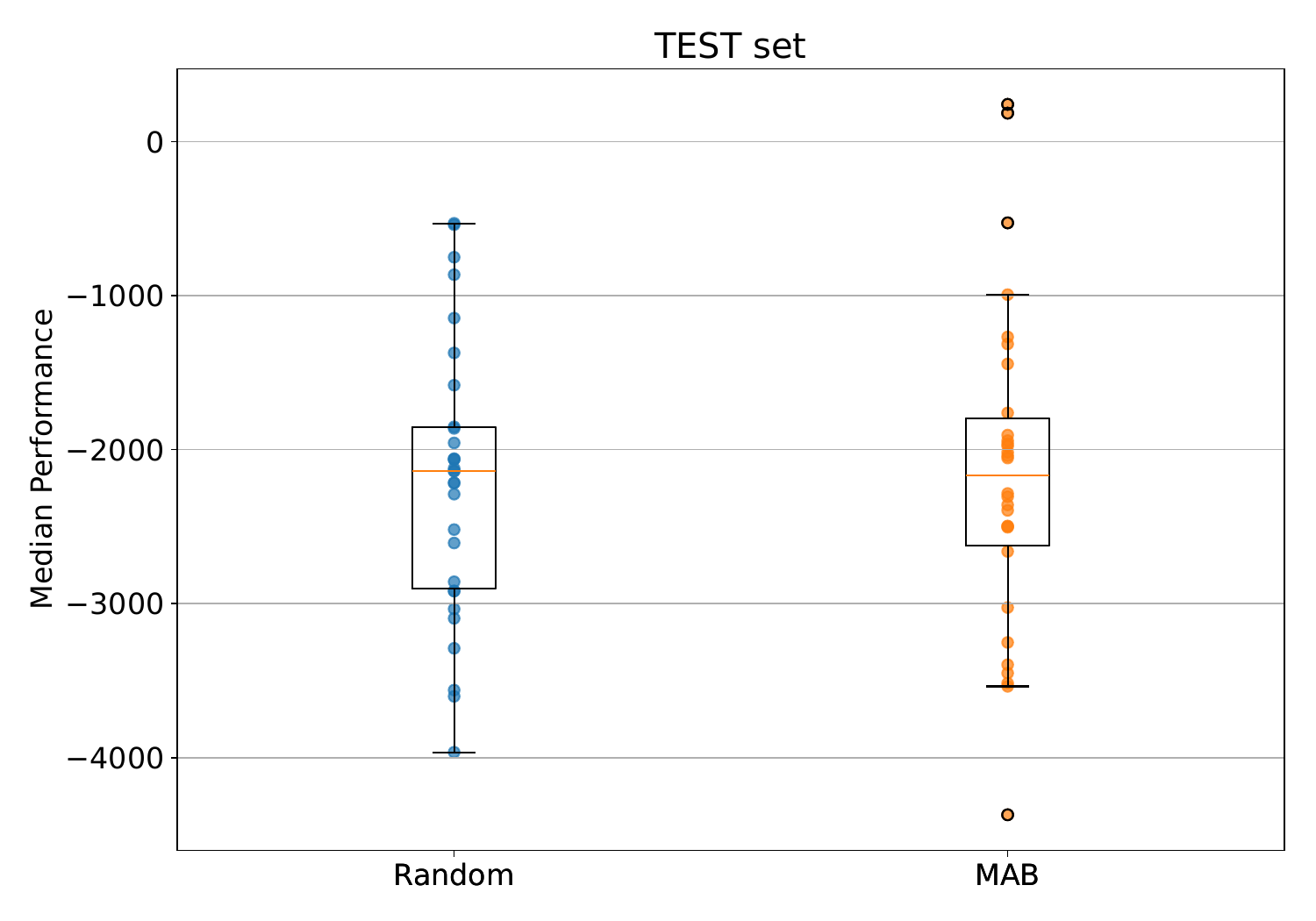}
 \caption{Ant}
 \Description{}
 \label{fig:Box_MAB_ant}
 \end{subfigure}
 \caption{Performance comparison between Random and Multi-Armed Bandit on the test morphologies.}
\end{figure}

\begin{figure}[ht!]
 \centering
 \begin{subfigure}{.45\textwidth}
 \centering
 \includegraphics[width=\linewidth]{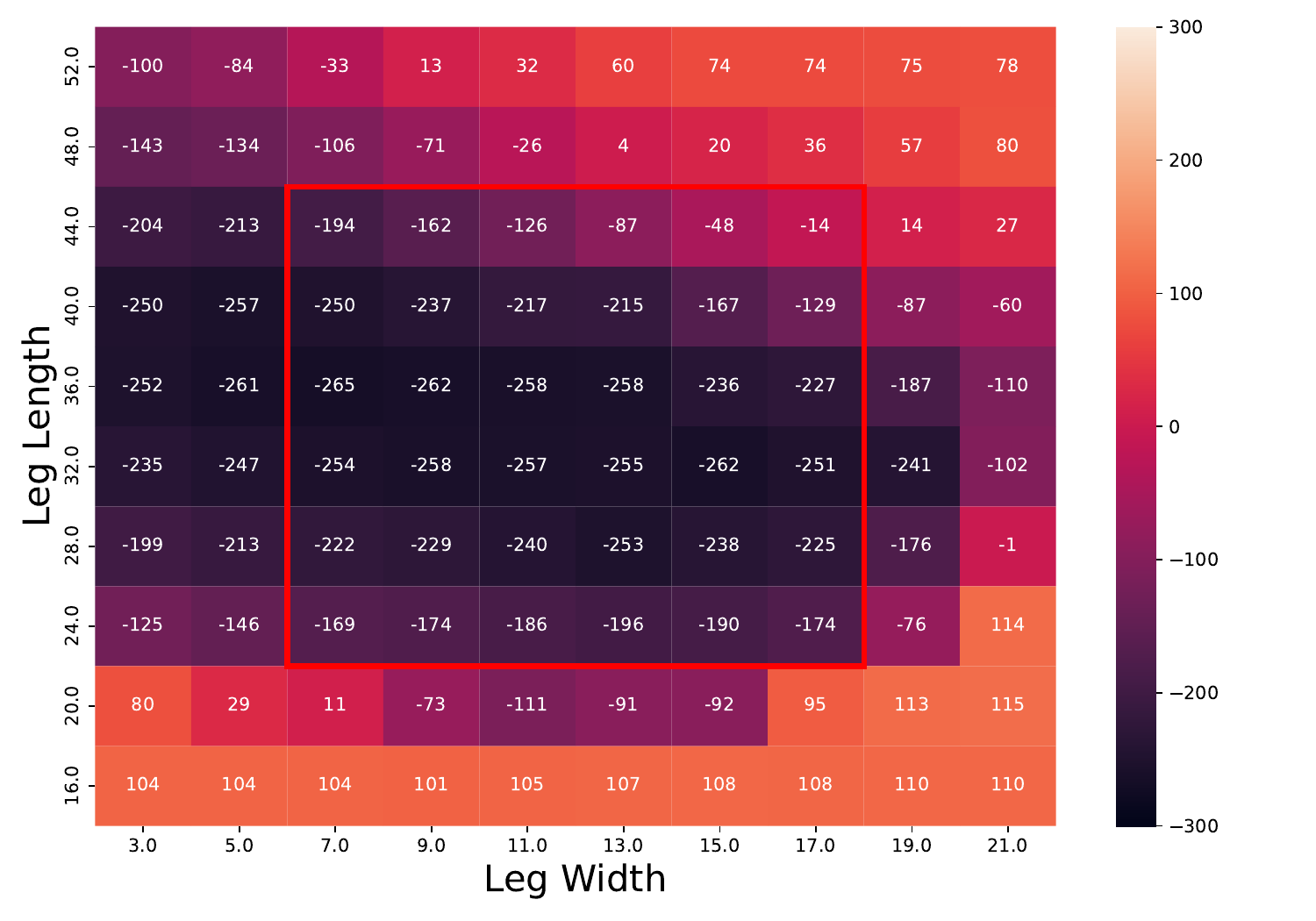}
 \caption{Random}
 \Description{}
 \label{fig:MAB_Walker_random}
 \end{subfigure}
 \hfill
 \begin{subfigure}{.45\textwidth}
 \centering
 \includegraphics[width=\linewidth]{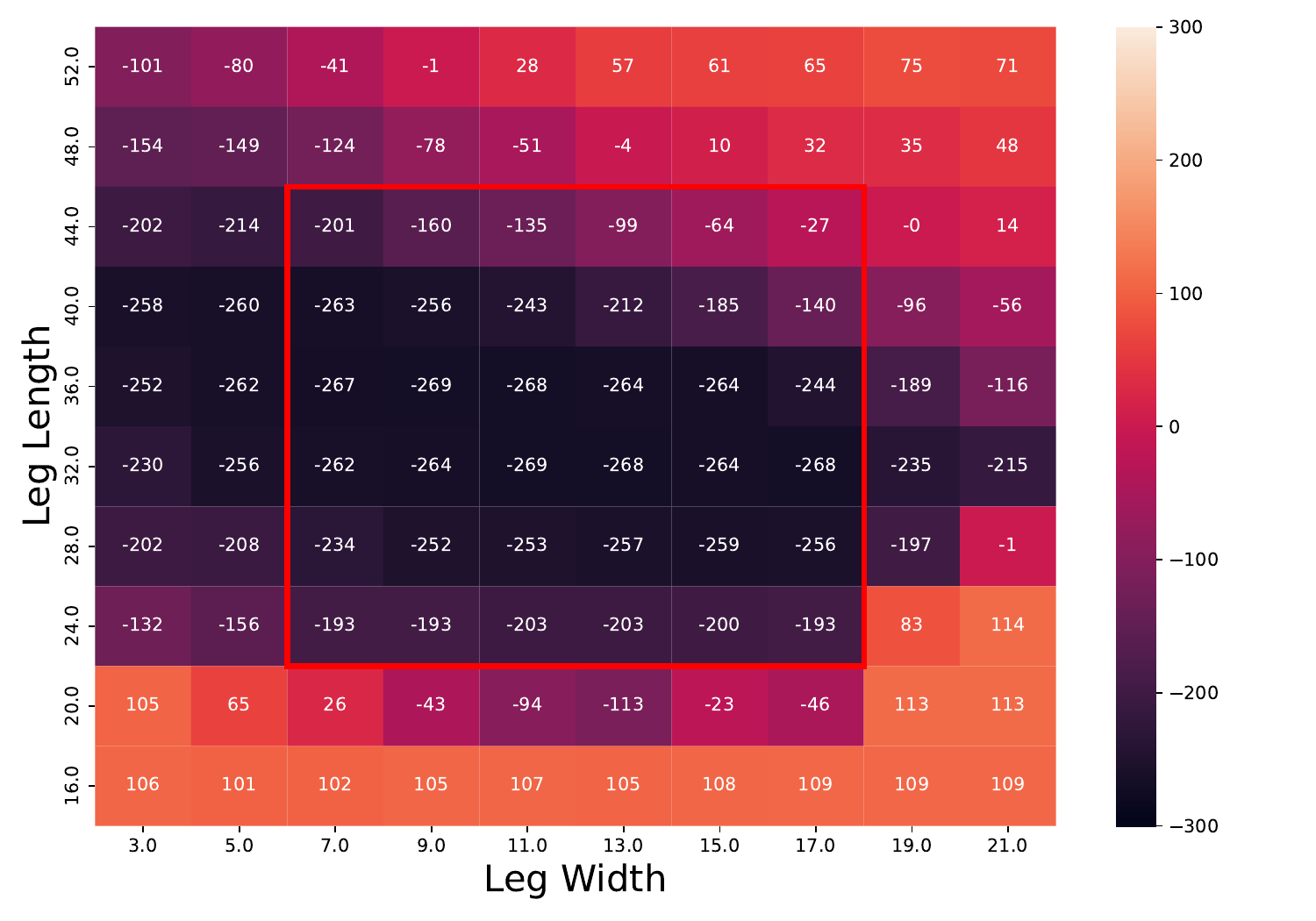}
 \caption{Multi-Armed Bandit}
 \Description{}
 \label{fig:MAB_Walker_MAB}
 \end{subfigure}
 \caption{Average performance across morphologies for Bipedal Walker.}
 \Description{}
 \label{fig:heatmap_MAB36_walker}
\end{figure}

\begin{figure}[ht!]
 \centering
 \begin{subfigure}{.45\textwidth}
 \centering
 \includegraphics[width=\linewidth]{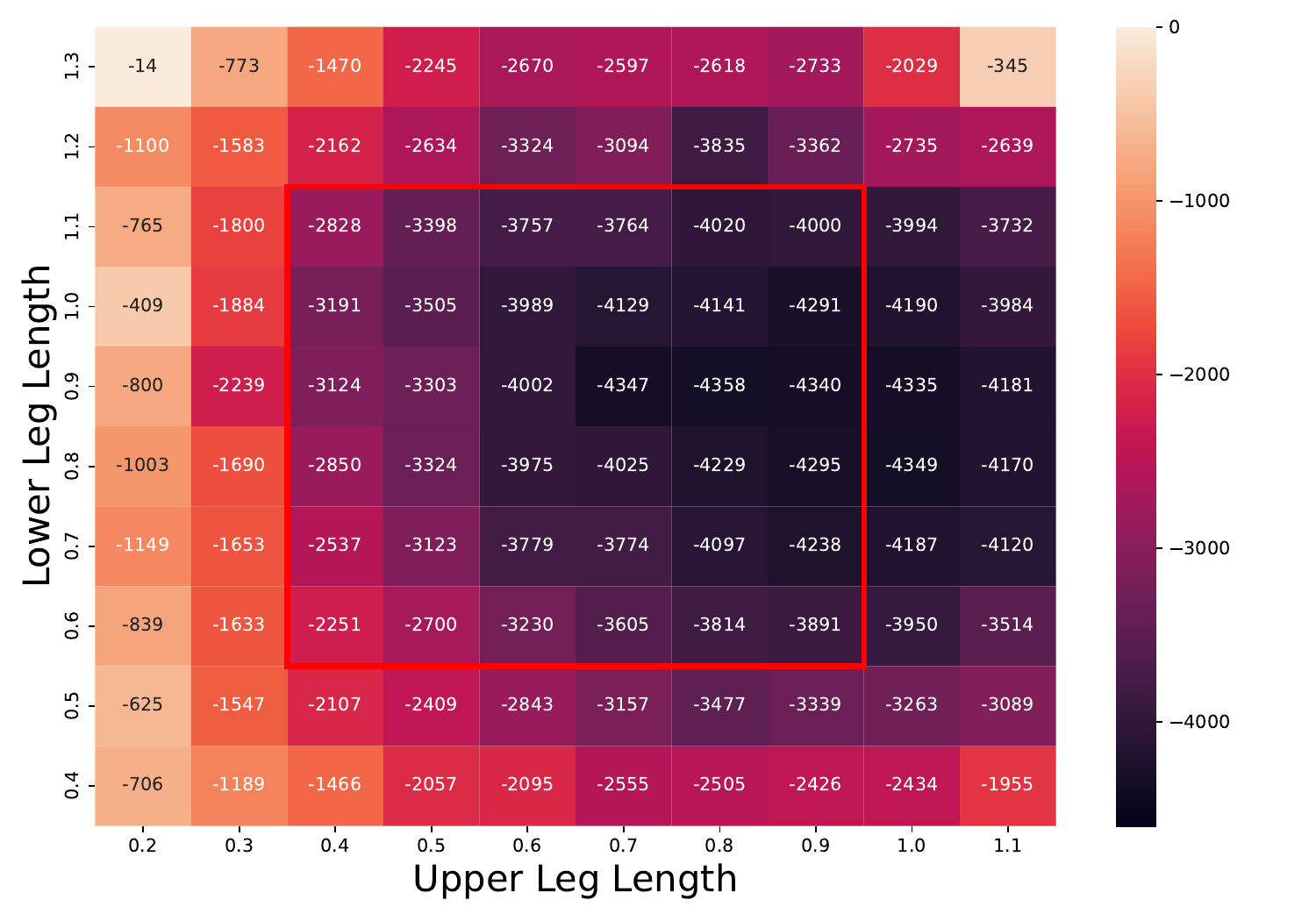}
 \caption{Random}
 \Description{}
 \label{fig:MAB_Ant_random}
 \end{subfigure}
 \hfill
 \begin{subfigure}{.45\textwidth}
 \centering
 \includegraphics[width=\linewidth]{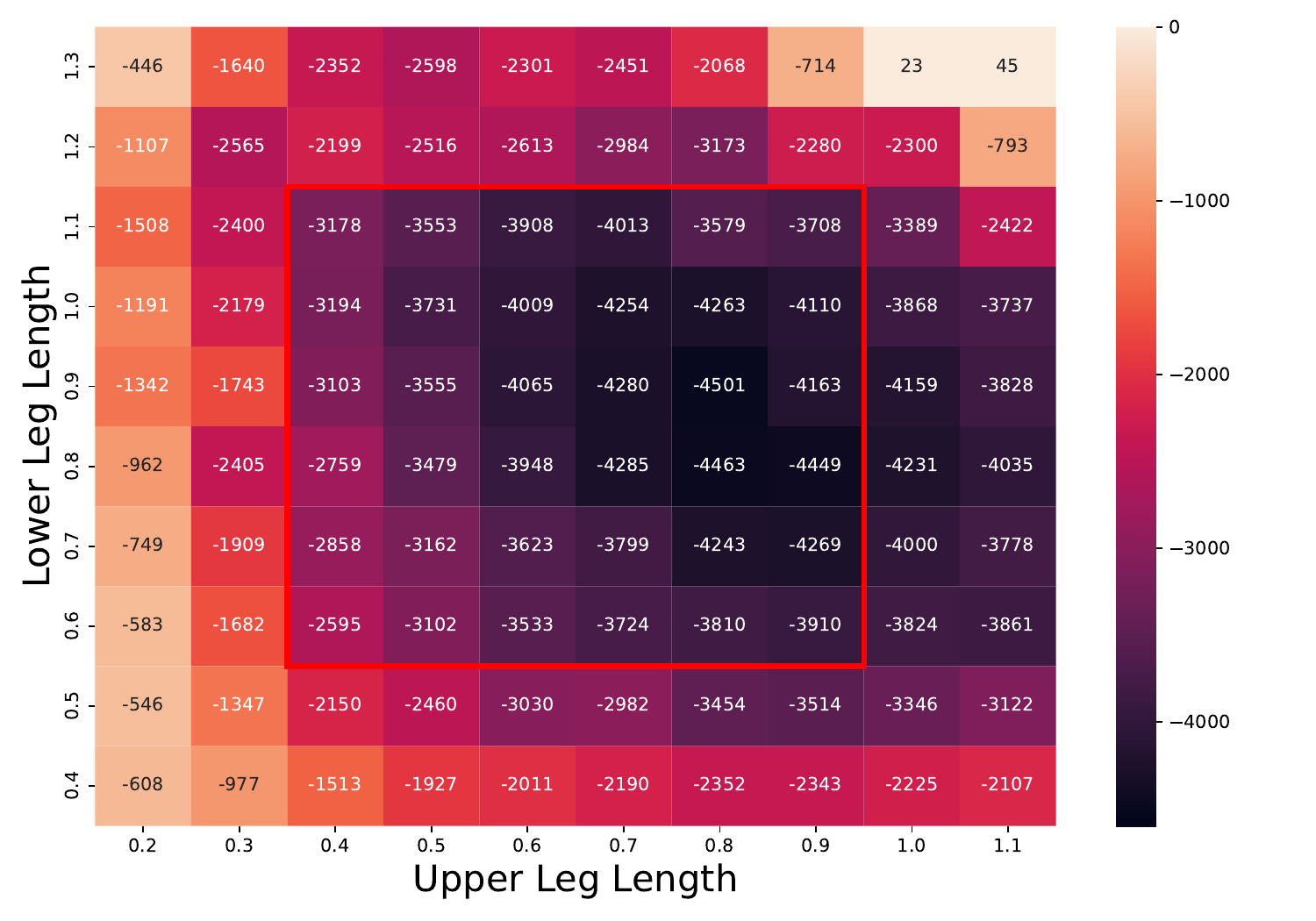}
 \caption{Multi-Armed Bandit}
 \Description{}
 \label{fig:MAB_Ant_MAB}
 \end{subfigure}
 \caption{Average performance across morphologies for Ant.}
 \label{fig:heatmap_MAB36_ant}
\end{figure}

Figures \ref{fig:Box_MAB_biped} and \ref{fig:Box_MAB_ant} illustrate a performance comparison between the Random and Multi-Armed Bandit (MAB) schedules on the testing set, specifically for the Bipedal Walker and Ant tasks, respectively. Figure \ref{fig:heatmap_MAB36_walker} and \ref{fig:heatmap_MAB36_ant} presents, respectively for Bipedal Walker and Ant, heatmaps illustrating the average performances of ANNs (across all the morphologies) trained using two different scheduling strategies: Random and MAB. Each cell in the heatmaps corresponds to a specific morphology. This visualization incorporates morphologies from both the training (inside the red square) and testing sets. The shading of each cell indicates the average performance of the ANN controller on that particular morphology, with darker shades indicating better scores. These averages are computed over all controllers obtained from 30 runs.

\subsection{Morphology frequency}
\begin{figure}[ht!]
\centering
 \begin{subfigure}{.45\textwidth}
 \centering
 \includegraphics[width=\linewidth]{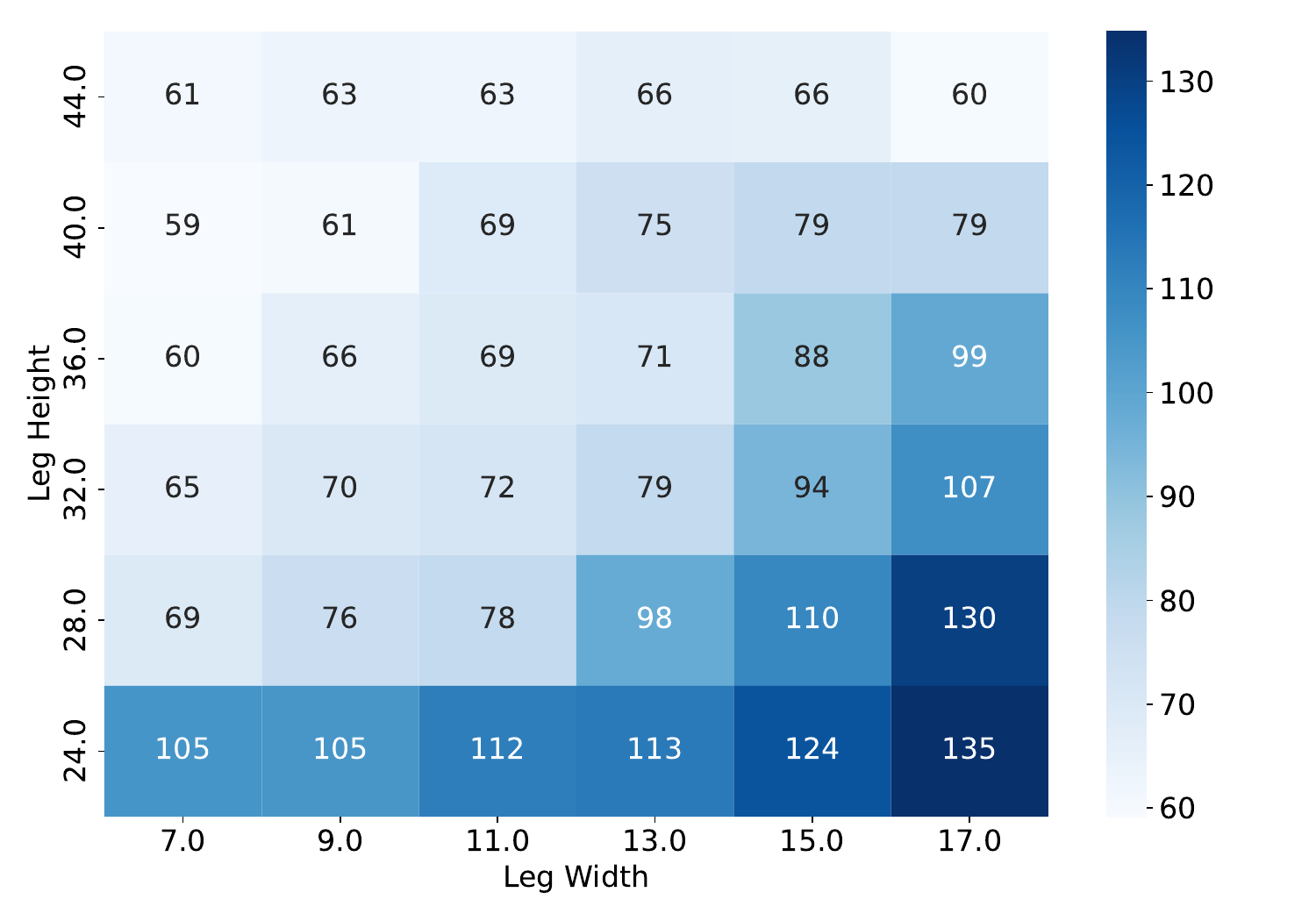}
 \caption{Bipedal Walker}
 \Description{}
 \label{fig:used_env36_biped}
 \end{subfigure}
 \hfill
 \begin{subfigure}{.45\textwidth}
 \centering
 \includegraphics[width=\linewidth]{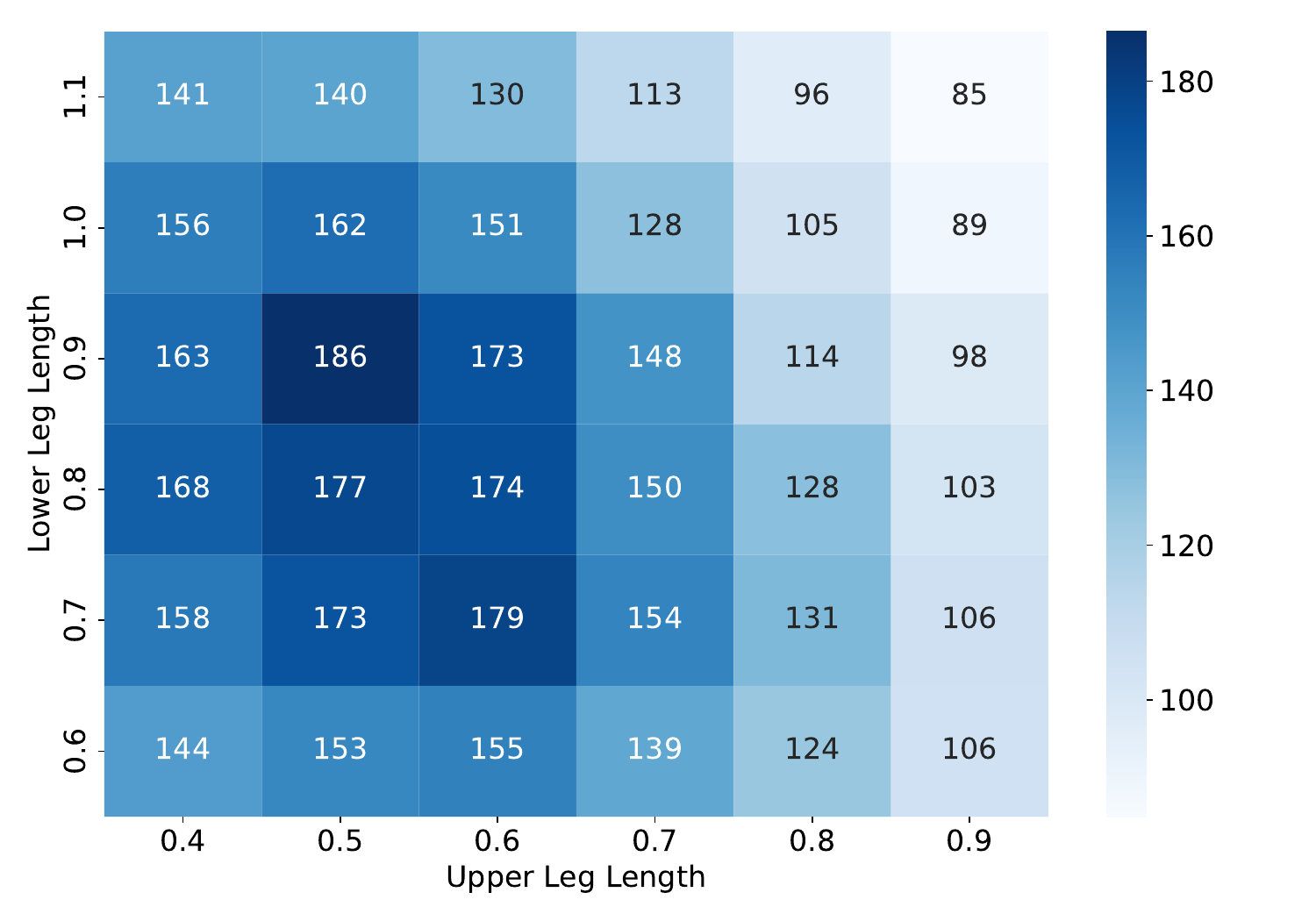}
 \caption{Ant}
 \Description{}
 \label{fig:used_env36_ant}
 \end{subfigure}
 \caption{Frequency of each morphology's selection during training.}
\end{figure}

Figures \ref{fig:used_env36_biped} and \ref{fig:used_env36_ant} display heatmaps showing the frequency of each morphology's selection during training for the Bipedal Walker and Ant tasks, respectively. Each cell in the heatmap represents a unique morphology, and the numerical value within the cell indicates the frequency with which that morphology was selected during training. This frequency is computed as the average occurrence across 30 independent training runs.

